\pdfoutput = 1
\documentclass{article}

\usepackage{PRIMEarxiv}

\usepackage[utf8]{inputenc} 
\usepackage[T1]{fontenc}    
\usepackage{hyperref}       
\usepackage{url}            
\usepackage{booktabs}       
\usepackage{amsfonts}       
\usepackage{nicefrac}       
\usepackage{microtype}      
\usepackage{lipsum}
\usepackage{fancyhdr}       
\usepackage{graphicx}       
\graphicspath{{media/}}     
\usepackage{wrapfig}
\usepackage{lipsum}
\usepackage{amssymb}
\usepackage{pifont}
\usepackage{caption}
\usepackage{amsmath}
\usepackage[normalem]{ulem}
\usepackage{fancyvrb}
\useunder{\uline}{\ul}{}
\usepackage{natbib}
\usepackage{multirow}
\usepackage{subcaption}

\usepackage{xcolor}
\usepackage[most,listings]{tcolorbox} 
\usepackage{fontawesome5}

\definecolor{sysIron}{HTML}{454545} 
\definecolor{sysBack}{HTML}{F9F9F9} 

\newtcblisting{systembox}[1][]{
  enhanced,
  breakable,                
  listing only,
  listing options={
    basicstyle=\ttfamily\small,
    breaklines=true,
    columns=fullflexible,
    keepspaces=true,
    literate={•}{\textbullet}1 {—}{---}1 {’}{{'}}1,
  },
  colback=sysBack,
  colframe=sysIron,
  coltitle=white,
  title={\textbf{\faCog\ System Prompt}},
  fonttitle=\bfseries\sffamily,
  attach boxed title to top left={xshift=0mm, yshift=0mm},
  boxed title style={
    colback=sysIron,
    outer arc=0pt,
    arc=0pt,
    boxrule=0pt,
  },
  sharp corners=northwest,
  rounded corners=southeast,
  rounded corners=northeast,
  rounded corners=southwest,
  arc=3mm,
  boxrule=0.5mm,
  left=4mm, right=4mm, top=4mm, bottom=4mm,
  toptitle=2mm, bottomtitle=2mm,
  drop shadow,
  #1
}

\RequirePackage{algorithmic}
\RequirePackage{algorithm}
\renewcommand{\algorithmicrequire}{\textbf{Input:}}
\renewcommand{\algorithmicensure}{\textbf{Output:}}
\renewcommand{\algorithmiccomment}[1]{\hfill $\triangleright$ \textit{#1}}

\title{Implicit Turn-Wise Policy Optimization for Proactive User-LLM Interaction}

\author{
  Haoyu Wang\textsuperscript{1}\footnotemark[1]\hspace{0.4em},
  Yuxin Chen\textsuperscript{2},
  Liang Luo\textsuperscript{2},
  Buyun Zhang\textsuperscript{2}, 
  Ellie Dingqiao Wen\textsuperscript{2},
  Pan Li\textsuperscript{1},\footnotemark[1]\hspace{0.4em} \\
  {\textsuperscript{1} Georgia Institute of Technology
  \textsuperscript{2} Meta AI}
}

\newcommand{\proj}{ITPO}

\definecolor{myorange}{rgb}{1, 0.5, 0}

\begin{document}

\maketitle

\renewcommand{\thefootnote}{\fnsymbol{footnote}}
\footnotetext[1]{haoyu.wang@gatech.edu, panli@gatech.edu.}

\begin{abstract}
Multi-turn human-AI collaboration is fundamental to deploying interactive services such as adaptive tutoring, conversational recommendation, and professional consultation. However, optimizing these interactions via reinforcement learning is hindered by the sparsity of verifiable intermediate rewards and the high stochasticity of user responses. 
To address these challenges, we introduce Implicit Turn-wise Policy Optimization (ITPO). ITPO leverages an implicit process reward model to derive fine-grained, turn-wise process rewards from sparse outcome signals. Unlike volatile token-level rewards, these turn-level signals exhibit superior robustness and may utilize a normalization mechanism to further enhance training stability. 
We evaluate ITPO across three representative multi-turn collaborative tasks: math tutoring, document writing, and medical recommendation. Empirical results demonstrate that ITPO, when combined with PPO, GRPO, or RLOO, consistently achieves improved convergence than existing baselines. Elaborate trajectory analysis confirms that ITPO infers turn-wise preferences that are semantically aligned with human judgment. Code is publicly available at \url{https://github.com/Graph-COM/ITPO}.
\end{abstract}

\section{Introduction}
Human-AI collaboration is revolutionizing the societal landscape~\cite{ziems2024can, li2023camel}, ranging from enhancing personalization in recommendation systems and education~\cite{zhu2024collaborative, zhao2024recommender} to the democratization of specialized expertise in healthcare and law~\cite{peng2023study, liu2023chatcounselor}.
Realizing this potential necessitates a paradigm shift in large language models (LLMs): moving beyond reactive instruction following toward multi-turn interactions, where agents must proactively resolve ambiguity~\cite{wu2025collabllm}, decompose complex goals~\cite{wei2022chain, jaech2024openai}, and iteratively refine their outputs~\cite{chen2024large}.
However, existing LLMs suffer from performance degradation in long-horizon interactions~\cite{laban2025llms}, making the effective alignment for multi-turn User-LLM collaboration via reinforcement learning (RL) an urgent yet open problem.

The core challenge lies in the supervision sparsity. As outcome rewards typically materialize until conversation concludes, relying only on these delayed signals often leads to poor sample efficiency and spurious solutions~\cite{sutton1998reinforcement}. Synthesizing fine-grained dense process rewards offers a potential solution, while existing approaches such as value estimation~\cite{konda1999actor, schulman2017proximal} or process reward models (PRMs)~\cite{ng1999policy} fall short under the multi-turn collaborative setting.
On one hand, training value models to estimate process rewards is empirically difficult~\cite{yuan2024free, guo2025deepseek, ahmadian2024back}, particularly given the high variance of user dynamics that hinders convergence.
On the other hand, current PRM pipelines lack scalability for multi-turn online RL~\cite{gao2023scaling}, relying either on labor-intensive human annotation (\textit{e.g.,} $800,000$ labels for math problems~\cite{lightman2023let}), or Monte Carlo roll-outs with 8-10 times more sample complexity~\cite{wang2024math, kazemnejad2024vineppo}.
Moreover, unlike structured reasoning or tool-execution tasks with verifiable step-level labels~\cite{zeng2025reinforcing, zhao2025mua, uesato2022solving}, defining the quality of an intermediate conversational turn is often ambiguous, requiring costly task-specific rubric design~\cite{yu2025sotopia}. As to off-the-shelf LLM judges, they introduce prohibitive latency for online RL optimization~\cite{yu2025sotopia, lee2023rlaif} and often suffer from evaluation bias~\cite{zheng2023judging}.

\begin{figure*}[t]
    \centering
    \includegraphics[width=0.96\linewidth]{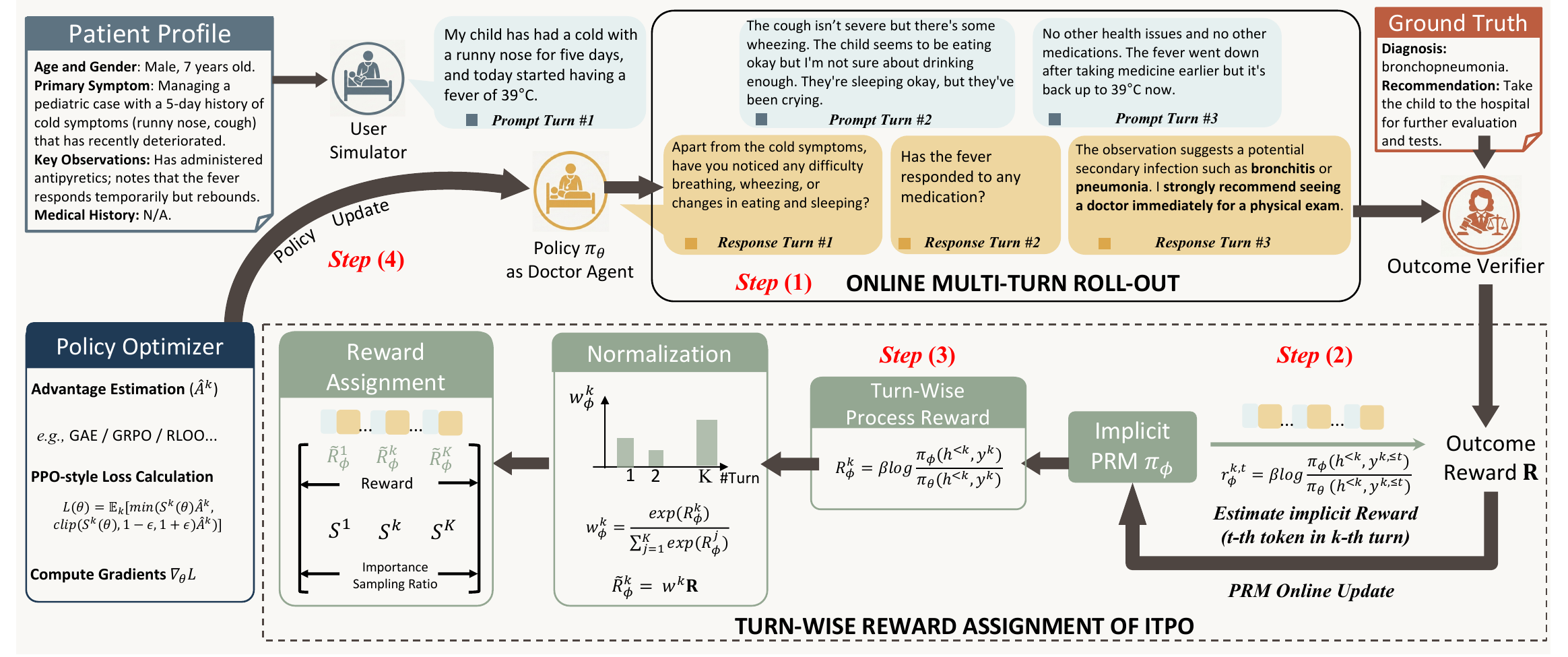}
    \vspace{-0.2cm}
    \caption{The overall framework of \proj for User-LLM interactions with a real case example in Medical Recommendation. The optimization loop consists: of (1) Multi-turn online roll-out, (2) implicit PRM update with outcome rewards and token-level process reward estimation, (3) Turn-wise process reward aggregation and normalization, (4) Advantage estimation and policy optimization.}
    \label{fig:framework}
    \vspace{-0.cm}
\end{figure*}

Recently, implicit process reward models (implicit PRMs) ~\cite{cui2025process,yuan2024free, rafailov2024r} have emerged as a scalable alternative, deriving dense token-level process rewards in single-turn tasks directly from outcome rewards without manual annotation or auxiliary sampling. However, this token-level granularity often induces high variance and overfitting~\cite{razin2025your}, while lacking semantic interpretability due to its excessive resolution. In contrast, within multi-turn User-LLM interactions, each turn serves as a natural atomic unit for semantic planning. This level of granularity is inherently interpretable, which may facilitate monitoring by domain experts and practitioners and algorithmically reduce variance by aggregating signals across all tokens within a turn. Motivated by these insights, we propose Implicit Turn-wise Policy Optimization (\proj), with our closed-loop online optimization framework illustrated in Figure~\ref{fig:framework}. Crucially, we introduce a normalization mechanism to calibrate the learned process rewards, thereby further enhancing training stability. \proj integrates seamlessly with various advantage functions, such as PPO~\cite{schulman2015high}, GRPO~\cite{shao2024deepseekmath}, and RLOO~\cite{ahmadian2024back, kool2019buy}, to facilitate PPO-style~\cite{schulman2017proximal} policy updates.

We evaluate \proj on three representative multi-turn collaborative tasks, namely Math Tutoring, Document Writing and Medical Recommendation. 
Analysis on training dynamics confirms that \proj learns stable and interpretable turn-wise preference, closely aligning with human judges. When integrated with PPO~\cite{schulman2017proximal}, RLOO~\cite{ahmadian2024back, kool2019buy} or GRPO~\cite{shao2024deepseekmath} advantage estimators, the turn-wise rewards provided by \proj consistently outperform sparse outcome-based baselines as well as reward shaping methods such as PRIME~\cite{cui2025process} and LLMs-as-a-Judge~\cite{yu2025sotopia}.

\section{Related Works}

\subsection{LLM Alignment Towards Multi-Turn Dynamics}
While existing post-training based on supervised fine-tuning~\cite{longpre2023flan} and RL~\cite{schulman2017proximal,rafailov2023direct,shao2024deepseekmath} prove effective for reactive instruction following~\cite{ouyang2022training}, the models struggle in multi-turn interactions~\cite{laban2025llms}. We categorize previous efforts to mitigate this gap according to two distinct application scenarios: the reasoning and the interaction scenarios.

\textbf{Reasoning Dynamics: Goal Decomposition and Tool Use.} \\The multi-step reasoning dynamics involve complex task decomposition, ranging from advanced problem-solving~\cite{guo2025deepseek, ouyang2023structured}, long-horizon gaming~\cite{wang2025vagen, wang2025ragen}, to agentic tool use~\cite{jin2025search, gao2023pal, huang2025visualtoolagent, pan2023logic}.
Trajectory-level standard PPO optimization~\cite{schulman2017proximal} serves as fundamental baselines~\cite{wang2025vagen, guo2025deepseek, zhou2025reinforced}. To address the reward sparsity~\cite{sutton1998reinforcement}, ~\cite{lightman2023let, uesato2022solving} train process reward models (PRMs) with up to 800k human annotated labels to evaluate intermediate math reasoning steps, yielding significant performance gains, while ~\cite{wang2024math}
construct step-level labels via Monte-Carlo sampling. 
In specific tasks, ~\cite{feng2025group} perform baseline estimation by clustering identical or similar states in finite environments, ~\cite{zeng2025reinforcing} derive turn-level rewards from verifiable tool-use labels. Crucially, these successes rely on the verifiability of intermediate steps.

\begin{figure*}[t]
  \centering
  \includegraphics[width=0.96\linewidth]{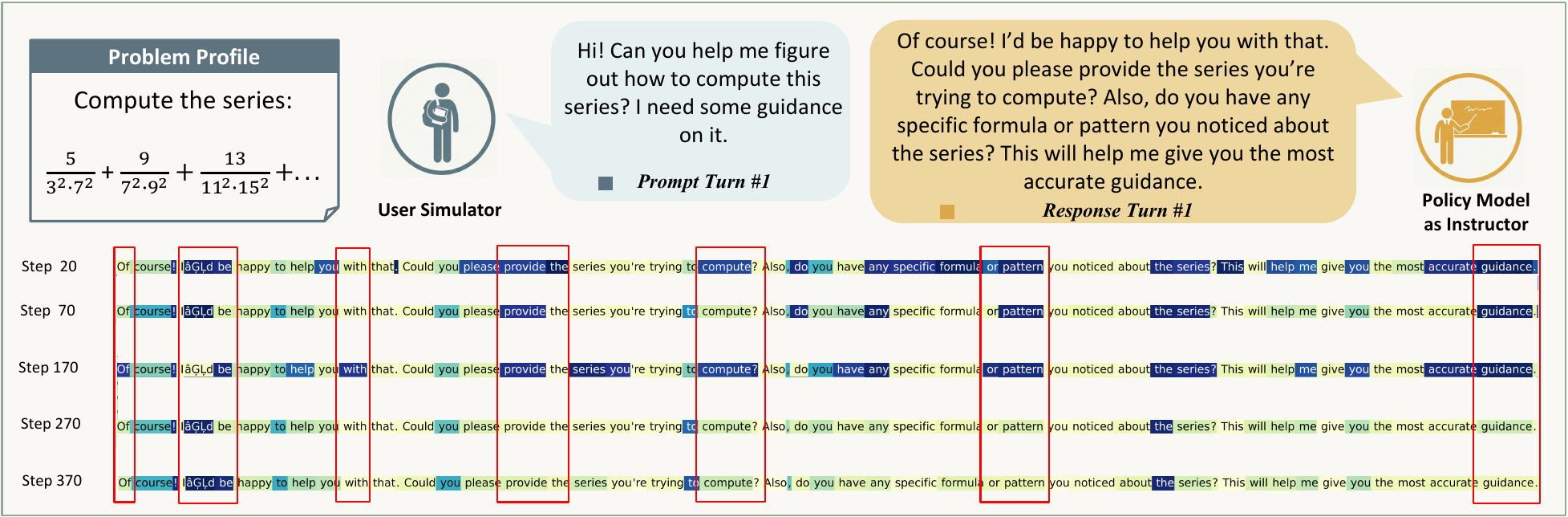}
  \caption{Dynamics of the relative implicit rewards on a real Math Tutoring case. Turn-wise rewards of Response Turn \#1 assigned by \proj consistently rank the highest from step $20$, meanwhile reflecting the importance of resolving ambiguity early in the conversation. By contrast, the token-level rewards show high variance, with the signals for identical tokens fluctuating across training steps (red boxes).}
    \label{fig:method}
    \vspace{-0.3cm}
\end{figure*}

\textbf{Interaction Dynamics: Dialogue and Collaboration.}
\\The interaction dynamics focus on collaborative multi-turn tasks involving human users or agent partners, spanning social science simulation~\cite{yu2025sotopia, wang2024sotopia}, recommendation systems~\cite{wan2025enhancing, bao2023tallrec}, and collaborative assistance~\cite{wu2025collabllm, sun2025training}. 
~\cite{wu2025collabllm} employ vanilla PPO with forward sampling for reward estimation in multi-turn collaborative tasks. Approaches that define fine-grained turn-wise rewards have shown effectiveness. For instance,  ~\cite{wan2025enhancing} model categorical user belief states within a POMDP framework for simplified recommendation, while Sotopia-RL~\cite{yu2025sotopia} leverage task-specific rigid rubrics and LLM-as-a-Judge to assign semantic turn-level rewards. However, these methods depend on restrictive task-specific priors. There is a critical need for scalable and automated algorithms to produce turn-wise rewards.
\\For more related works, see Appendix.~\ref{app:more_related_works}.

\section{Methodology}

We present the methodology by first formalizing the problem of the multi-turn collaborative task and second the preliminary fundamentals of implicit PRM. Subsequently, we introduce \proj, describing how it derives robust turn-wise rewards to enable scalable online policy optimization.

\subsection{Problem Formulation}
We formulate the multi-turn collaborative task as a Partially Observable Markov Decision Process (POMDP)~\cite{sondik1971optimal, aastrom1965optimal}, defined by the tuple $\langle \mathcal{S}, \mathcal{Y}, \mathcal{T}, \mathbf{R}, \gamma \rangle$. The state space $\mathcal{S} = \mathcal{H} \times \mathcal{G}$ comprises the observable conversation history $h^k \in \mathcal{H}$ and the user's latent goal $g \in \mathcal{G}$, which is unobservable to the agent. The action space corresponds to the set of model responses $\mathcal{Y}$. At turn $k$, the agent policy $\pi_\theta$ generates a response $y^k \in \mathcal{Y}$ conditioned on $h^k$. The transition dynamics $\mathcal{T}$ are governed by the distribution of user's responses $P_U(x^{k+1} | h^k, y^k, g)$, where the user query $x^{k+1}$ serves as the observation for the next turn.
The interaction terminates after $K$ turns, yielding a trajectory $\tau = [(x^1, y^1), \dots, (x^K, y^K)]$ with a sparse outcome reward $\mathbf{R}(\tau) \in [0, 1]$\footnote{\fontsize{7.5pt}{9.4pt} Without loss of generality, while we adopt the probabilistic interval $[0,1]$ to ground the derivation, the proposed attribution mechanism remains valid for any bounded reward scale $\mathbf{R}\in[a,b]$ via linear scaling. Furthermore, \proj can accommodate multiple reward signals through linear aggregation (i.e., $\mathbf{R}=\sum_m \mathbf{R}^{(m)}$).}, representing the quality of the collaboration. Consequently, the core objective is to maximize the expected outcome reward $\mathbb{E}[\mathbf{R}]$. Given the finite horizon of dialogue tasks, discount factor $\gamma$ is set to $1$. Throughout this paper, we prompt an LLM-based user simulator to approximate $P_U$~\cite{dou2025simulatorarena}.

Optimizing the objective above presents two challenges. First, the distribution $P_U$ of the observation $x^k$ given the latent goal $g$ could have high variance: users with identical goals may formulate diverse queries or distinct interaction styles. The agent needs to implicitly reason about $g$ under high uncertainty. Second, accurate value estimation is challenging. The sparse reward $\mathbf{R}(\tau)$, compounded by the high variance in user dynamics, renders the estimation of $V_\psi(h^k) = \mathbb{E}_{\tau} [\mathbf{R}(\tau) | h^k]$ sample-inefficient. 
A potential solution is reward shaping~\cite{ng1999policy}, which decomposes the outcome $\mathbf{R}(\tau)$ into a sequence of dense, proxy rewards to provide fine-grained feedback.

\subsection{Preliminary: Implicit Process Reward Model}
Implicit PRMs~\cite{yuan2024free, cui2025process} parameterize dense, token-level process rewards via generative models, enabling online update using solely outcome rewards. Let $y^{k,t}$ denote the $t$-th token of response $y^k$ at turn $k$, the corresponding token-level process reward is defined as the scaled log-likelihood ratio between an implicit PRM $\pi_\phi$ and a fixed reference model $\pi_{\text{ref}}$:
\begin{equation}
    \label{eq:reward_def}
    r_\phi(y^{k,t} \mid h^k, x^k, y^{k, <t}) = \beta \log \frac{\pi_\phi(y^{k,t} \mid h^k, x^k, y^{k, <t})}{\pi_{\text{ref}}(y^{k,t} \mid h^k, x^k, y^{k, <t})},
\end{equation}
where $\beta$ is a scaling coefficient, $h^k$ is the history before turn-$k$, $x^k$ is the user query, and $y^{k, <t}$ denotes the prefix. The trajectory-level implicit reward $\mathcal{R}_\phi(\tau)$ aggregates these token-level rewards over the sequence:
\begin{equation}
    \mathcal{R}_\phi(\tau) = \sum_{k=1}^K \sum_{t=1}^{|y^k|} r_\phi(y^{k,t} \mid h^k, x^k, y^{k, <t}).
\end{equation}
Optimization aligns $\mathcal{R}_\phi(\tau)$ with the ground-truth outcome $\mathbf{R}(\tau)$ via binary cross-entropy:
\begin{equation}
    \label{eq:prm_loss}
    \mathcal{L}_{\text{PRM}}(\phi) = - \mathbb{E}_{(\tau, \mathbf{R}) \sim \mathcal{D}} \left[ \mathbf{R} \log \sigma \left( \mathcal{R}_\phi(\tau) \right) + \\ (1-\mathbf{R}) \log \left( 1 - \sigma \left( \mathcal{R}_\phi(\tau) \right) \right) \right],
\end{equation}
where $\sigma$ denotes sigmoid function. Invoking \textbf{Proposition 3.1} from \cite{cui2025process}, the partial accumulation functions as an implicit Q-value estimator, $Q_\phi(h^k, x^k, y^{k, \le t}) \approx \sum_{j=1}^{k-1}\sum_{p=1}^{|y^j|} r_\phi(y^{j,p} \mid \cdot) + \sum_{i=1}^t r_\phi(y^{k,i} \mid \cdot)$, establishing $r_\phi$ as the token-wise implicit value increment.

\begin{wrapfigure}{r}{0.47\textwidth}
    \centering
    \vspace{-0.5cm}
    \includegraphics[width=0.47\textwidth]{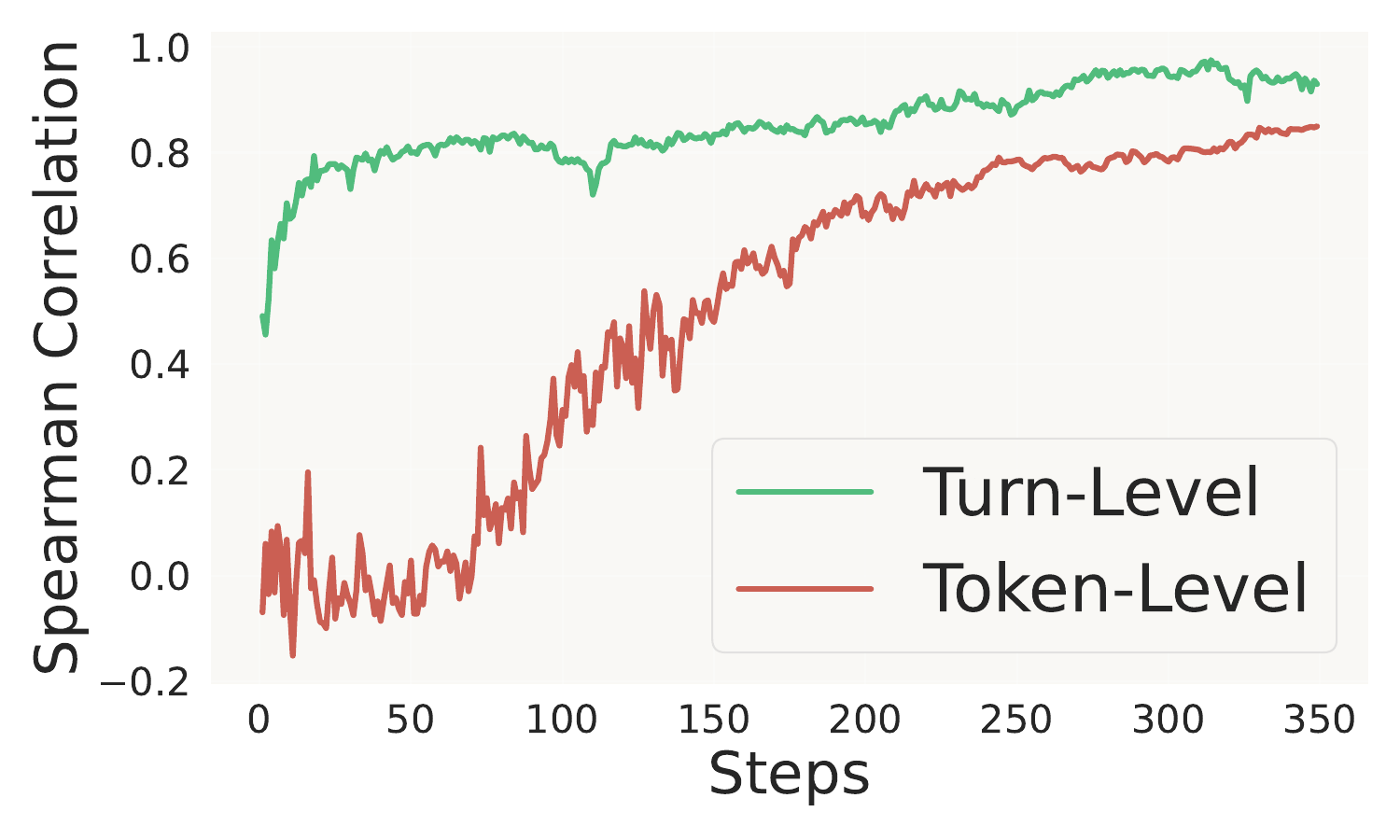}
    \vspace{-0.3cm}
    \caption{Spearman correlation of the token-level and turn-wise implicit rewards against their respective converged baselines (the averaged rewards over the final $70$ optimization steps) on Math Tutoring. Turn-wise preference stabilizes quickly, while token-level rankings show slower convergence due to optimization difficulty.}
    \label{fig:token_vs_turn}
    \vspace{-0.4cm}
\end{wrapfigure}

Standard token-level implicit PRMs face three limitations. (1) High variance for token-level reward. As visualized by the red boxes in Fig.~\ref{fig:method}, while the aggregated turn-wise score correctly identifies Response Turn \#1 as the pivotal turn after just $20$ updates, the internal token-level rewards remain chaotic along training steps. Also as shown in Fig.~\ref{fig:token_vs_turn}, turn-wise preferences stabilize rapidly, whereas token-level rewards exhibit significantly slower convergence.
(2) Semantic misalignment. As in the example, high rewards fluctuate arbitrarily across functional tokens (e.g., ``with'', ``please") during training. (3) Overfitting Issue. Theoretically the overly fine-grained supervision exacerbates the susceptibility to overfitting~\cite{razin2025your}.

\subsection{Implicit Turn-Wise Policy Optimization (\proj)}
\label{sec:method_itpo}
Given the preliminaries, we introduce \proj and its normalized version Norm-\proj. 

\textbf{ITPO} mitigates token-level noise by aggregating the token-level evidence, where the turn-wise implicit reward $\mathcal{R}_\phi^k$ is defined as the accumulation of token-level log-likelihood ratios within the $k$-th turn:
\begin{equation}
    \label{eq:itpo_score}
    \mathcal{R}_\phi^k = \sum_{t=1}^{|y^k|} r_\phi(y^{k,t} \mid h^k, x^k, y^{k, <t}),
\end{equation}
where $|y^k|$ denotes the length of the response at turn $k$. Physically, $\mathcal{R}_\phi^k$ represents the cumulative implicit advantage of the turn. \proj effectively mitigates the high variance and noise inherent in individual token-level estimations, as illustrated in Fig.~\ref{fig:token_vs_turn}.

\textbf{Norm-\proj}. While the learned trajectory-level implicit reward $\mathcal{R}_{\phi}(\tau)$ exhibits a positive correlation with the outcome reward $\mathbf{R}(\tau)$, the scale of such mapping suffer from fluctuation, even on fixed roll-out trajectories (See Fig.~\ref{fig:stability}). The instability poses two challenges: it introduces variance into policy updates and creates a non-stationary target that hinders value model convergence. These observations motivate Norm-\proj, which explicitly calibrates the reward scale to ensure optimization stability.

Norm-\proj utilizes the aggregated turn-wise scores to redistribute the global outcome reward $\mathbf{R}$. We employ a Softmax function to compute the contribution weight $w_k$ for each turn $k$:
\begin{equation}
    \label{eq:norm_weights}
   w_{\phi}^k = \text{Softmax}(\mathcal{R}_\phi^k / \eta) = \frac{\exp(\mathcal{R}_\phi^k / \eta)}{\sum_{j=1}^K \exp(\mathcal{R}_\phi^j / \eta)},
\end{equation}
where $\eta$ is a temperature hyperparameter. The normalized reward for turn $k$ is then computed as a fraction of the global outcome:
\begin{equation}
    \label{eq:norm_reward}
    \tilde{\mathcal{R}}_{\phi}^k = w_{\phi}^k \cdot \mathbf{R},
\end{equation}

By doing normalization, Norm-\proj not only provides inter-turn preference but also enforces scale consistency between the implicit reward and the outcome $\mathbf{R}$, which prevents value drift and stabilizes the training dynamics. Notably, this formulation generalizes two fundamental strategies: as $\eta \to 0$, the reward is assigned solely to the most pivotal turn, a strong baseline in social simulation~\cite{yu2025sotopia}; conversely, as $\eta \to \infty$, the distribution becomes uniform, corresponding to a mean-contribution baseline where every step shares the contribution equally.

\textbf{Bayesian Interpretation of Norm-\proj.}
The redistribution mechanism can be grounded in Bayesian inference over a latent pivotal turn $Z \in \{1, \dots, K\}$. Adopting a uniform prior $P(Z)=1/K$ and modeling the likelihood of the implicit evidence $\mathcal{R}_\phi^k$ as a Boltzmann distribution $P(\mathcal{E} \mid Z=k) \propto \exp(\mathcal{R}_\phi^k / \lambda)$, the posterior probability of turn $k$ being the decisive factor is derived via Bayes' theorem:
\begin{equation}
    P(Z=k \mid \mathcal{E}) = \frac{\exp(\mathcal{R}_\phi^k / \eta)}{\sum_{j=1}^K \exp(\mathcal{R}_\phi(y_j) / \eta)},
\end{equation}
which equals to weight $w^k$. Consequently, the assigned reward $\tilde{\mathcal{R}}_{k}$ represents the expected credit attributed to turn $k$ under the posterior belief.

Overall, \proj and Norm-\proj provide fine-grained turn-wise process rewards allocation that exhibits self-consistency, stability, and interpretability. Refer to Section~\ref{sec:turn_wise_allocation} for a comprehensive empirical validation.

\subsection{Policy Optimization}
\label{sec:policy_optimization}
The derived turn-wise rewards can be integrated with standard advantage estimators. For the $k$-th turn in the $i$-th sample, the advantage $\hat{A}^{i,k}$ is computed as follows:

\textbf{RLOO~\cite{ahmadian2024back}:} This method uses a leave-one-out baseline to reduce variance without a parametric value model. The advantage is estimated as $\hat{A}^{i,k}_{\text{RLOO}} = \tilde{\mathcal{R}}_{\phi}^{i,k} - \frac{1}{G-1} \sum_{j \neq i} \tilde{\mathcal{R}}_{\phi}^{j,k}$, where $G$ denotes the number of roll-out trajectories in the group sharing same prompts.

\textbf{GRPO~\cite{shao2024deepseekmath}:} To stabilize training, GRPO standardizes rewards based on group statistics. The advantage is computed as $\hat{A}^{i,k}_{\text{GRPO}} = (\tilde{\mathcal{R}}_{\phi}^{i,k} - \mu^k) / (\sigma^k + \epsilon)$, where $\mu^k$ and $\sigma^k$ represent the mean and standard deviation of the group rewards $\{\tilde{\mathcal{R}}_{\phi}^{j,k}\}_{j=1}^G$, respectively.

\textbf{Learned Value Model~\cite{schulman2015high}:} Following the PPO standard, this approach employs Generalized Advantage Estimation (GAE) with a separate critic network $V_\psi$. The advantage is defined as $\hat{A}^{i,k}_{\text{PPO}} = \sum_{l=0}^{K-k-1} (\gamma \lambda)^l \delta^{i, k+l}$, where the term $\delta^{i, t} = \tilde{\mathcal{R}}_{\phi}^{i,t} + \gamma V_\psi(h^{i, t+1}) - V_\psi(h^{i,t})$ represents the turn-level Temporal Difference (TD) error.

\textbf{Loss Function.} 
The policy $\pi_\theta$ is updated via PPO loss~\cite{schulman2017proximal}, with turn-level importance sampling ratio defined as $\rho^{i,k}(\theta) = \frac{\pi_\theta(y^{i,k} \mid h^{i,k})}{\pi_{\text{old}}(y^{i,k} \mid h^{i,k})}$. The objective is to maximize the clipped surrogate objective:
\begin{equation}
    \label{eq:ppo_loss}
    \mathcal{L}_{\text{PPO}}(\theta) = \mathbb{E} \left[ \sum_{k=1}^K \min \left( \rho^{i,k} \hat{A}^{i,k}, \ \text{clip}(\rho_{i,k}, 1-\epsilon, 1+\epsilon) \hat{A}^{i,k} \right) \right].
\end{equation}
This formulation shifts optimization from tokens to turns. By enforcing the sampling ratio on the entire turn, \proj preserves semantic coherence and avoids the disruption of joint probability dependencies caused by token-level clipping~\cite{zheng2025group}.

\section{Experiments}

\subsection{Settings and Implementation}

\subsubsection{Datasets and Metrics}
Experiments are conducted on three representative multi-turn collaborative use cases: education (math tutoring), content generation (document writing) and knowledge-intensive recommendation (medical recommendation)~\cite{tamkin2024clio,sun2025training, achiam2023gpt, touvron2023llama}. For details, see Appendix.~\ref{app:task_details}.

\textbf{Document Writing} is representative for iterative content generation. The core challenge is to ensure alignment with the user's latent intent through iterative refinement. Each instance is anchored by a ground-truth \textit{goal document} and a high-level summary. The user simulator is prompted with the summary, utilizing it to guide the LLM's drafting. Performance is quantified with \textbf{BLEU} score, measuring the semantic overlap between the generated draft and the goal. We adopt the protocol from \cite{wu2025collabllm}, utilizing a dataset of $500$ curated articles sourced from Medium.

\textbf{Math Tutoring} evaluates the capability in handling under-specified queries. In real-world tutoring, students often provide incomplete problem descriptions, requiring the tutor to actively solicit information through iterative interactions. In our setup, each instance consists of a math problem and its ground truth solution. The user simulator holds the full problem but reveals them partially and deliberately to the LLM. Task success is evaluated by the \textbf{Accuracy (ACC)} of the answer, as verified by an LLM judge. The dataset comprises $500$ problems from MATH~\cite{hendrycks2021measuring}, following the configuration in \cite{wu2025collabllm}.

\textbf{Medical Recommendation} targets knowledge-intensive recommendation. We model this as a diagnostic consultation where a patient presents symptoms to a doctor. The LLM performs symptom elicitation to reach a diagnosis and a recommendation rooted in expert knowledge. The user simulator is conditioned on a patient profile. Performance is assessed via a hierarchical \textbf{Score} assigned by an expert LLM judge. Score 0 is assigned for incorrect diagnoses; scores 1--3 for correct diagnoses paired with suboptimal or hallucinated recommendations; and scores 8--10 for accurate diagnoses. The dataset consists of $550$ samples from the MTMedDialog benchmark~\cite{feng2025doctoragent}.

\subsubsection{Baselines for Reward Shaping}
We compare \proj against five baselines with granularity from trajectory to token-level. Details in Appendix.~\ref{app:baseline_details}.

\textbf{Share along Trajectory:} 
    The standard outcome supervision baseline that underpins RLOO~\cite{ahmadian2024back} and GRPO~\cite{shao2024deepseekmath}. It broadcasts the final outcome reward $\mathbf{R}$ across all tokens in the trajectory. The method ignores the temporal structure of the interaction, treating every token as an equal contribution.

\textbf{Uniform Decomposition:}
    distributes the outcome reward uniformly with randomness. The weights are modeled with Dirichlet distribution $\mathbf{w} \sim \text{Dirichlet}(\mathbf{1})$, with turn-wise rewards defined as $R^k = w_k \cdot \mathbf{R}$. This provides a dense reward baseline where every interaction turn contributes a uniform and positive marginal gain towards the final outcome.

\textbf{Value Model:} 
    learns a value function $V_\psi$ to provide dense, token-level feedback~\cite{schulman2017proximal}. The reward signal is derived from the temporal difference (TD) error: $\delta^{k,t} = \gamma V_\phi(h^{<k}, y^{k, \le t}) - V_\phi(h^{<k}, y^{k, < t})$~\cite{schulman2015high}. $\gamma$ is set as $1$ due to the finite nature of the task. While this offers fine-grained granularity, training a critic network may suffer from high variance and slow convergence.

\textbf{LLM-as-Judge:} 
    leverages the semantic reasoning capabilities of an external judge, which is shown effective in social tasks~\cite{yu2025sotopia}. The judge is prompted to analyze the conversation and assign a normalized scalar weight $w^k$ to each turn resulting in a shaped reward $R^k = w^k \mathbf{R}$. Given the prohibitive cost and latency of API-based models in online RL, we employ Qwen2.5-14B-Instruct, which demonstrates comparable capability to GPT-4o-mini~\cite{achiam2023gpt} in the report~\cite{yang2025qwen2}.

\textbf{Implicit PRM:}~\cite{yuan2024free} formulated as $r_\phi^{k,t} = \beta \log \frac{\pi_\phi(\cdot)}{\pi_\text{ref}(\cdot)}$. While this method serves as the atomic evidence for \proj, we include it as a baseline to explicitly demonstrate the limitations of token-level supervision in multi-turn interactions, which often suffer from high variance and susceptible to overfitting~\cite{razin2025your}.

\subsubsection{Implementation}
\textbf{Backbone} We employ Qwen2.5-14B-Instruct as the user simulator. Main evaluations use Qwen2.5-3B-Instruct~\cite{yang2025qwen2} as policy. To verify the generalizability of our method, we provide additional results with Qwen3-4B~\cite{yang2025qwen3} and Qwen2.5-7B-Instruct~\cite{yang2025qwen2} as policy. The implicit PRM is initialized with the same parameters as the policy.
We perform full-parameter fine-tuning on the policy across all experiments. 

\textbf{Reward Design}
To discourage verbosity without compromising task performance, the optimization objective is formulated as the combination of the task-specific metric and a length regularization term. The objective is to maximize a score $S = \mathbf{R} - \gamma \times N$, where $\mathbf{R} \in [0,1]$ represents task-specific metric as outcome reward and $N$ is the token count of the model output, we set $\gamma$ as 5e-6 across all the tasks.

\textbf{Reward Model Training} If training PRM is required, both the baselines and \proj are trained with the outcome metric $\mathbf{R}$ as objective. The token amount regularization is explicitly calculated and assigned to individual turns.

\textbf{Policy Optimization} In all experiments, we employ the PPO loss defined in Equation.~\eqref{eq:ppo_loss}, with a KL penalty coefficient of 1e-3 by default. We evaluate three advantage estimators: RLOO~\cite{kool2019buy}, GRPO~\cite{shao2024deepseekmath}, and GAE~\cite{schulman2015high}, calculated using equations in Section.~\ref{sec:policy_optimization}. Details in Appendix.~\ref{app:itpo_implementation_details}.
\subsection{Main Results}


\begin{table*}[h]
\centering
\resizebox{0.94\textwidth}{!}{\begin{tabular}{cc|ccc|ccc|ccc}
\hline
\multirow{2}{*}{\begin{tabular}[c]{@{}c@{}}Advantage\\ Estimator\end{tabular}} & \multirow{2}{*}{\begin{tabular}[c]{@{}c@{}}Credit\\ Assignment\end{tabular}} & \multicolumn{3}{c|}{Math Tutoring} & \multicolumn{3}{c|}{Doc Writing} & \multicolumn{3}{c}{Medical Recommendation} \\ \cline{3-11} 
 &  & Accracy$\uparrow$ & Token(k)$\downarrow$ & Score$\uparrow$ & Bleu$\uparrow$ & Token(k)$\downarrow$ & Score$\uparrow$ & Diagnosis$\uparrow$ & Token(k)$\downarrow$ & Score$\uparrow$ \\ \hline
\multicolumn{2}{c|}{Vanilla} & 25.81 & 1.70 & 17.29 & 24.97 & 2.40 & 12.93 & 25.52 & 0.28 & 24.08 \\ \hline
\multirow{4}{*}{PPO} & VM Only & {\ul 26.81} & 1.36 & 20.06 & 40.57 & \textbf{0.84} & 36.37 & {\ul 58.41} & {\ul 0.72} & 54.77 \\
 & VM+PRIME & 22.63 & 1.54 & 14.92 & 41.87 & 2.42 & 29.77 & 52.07 & \textbf{0.48} & 49.73 \\
 & VM+ITPO (Ours) & 26.38 & {\ul 0.94} & {\ul 21.51} & {\ul 42.52} & 1.02 & {\ul 37.44} & 58.35 & 0.68 & {\ul 54.87} \\
 & VM+Norm-ITPO (Ours) & \textbf{27.15} & \textbf{0.74} & \textbf{23.44} & \textbf{45.16} & {\ul 0.96} & \textbf{40.12} & \textbf{62.44} & 0.60 & \textbf{59.44} \\ \hline
\multirow{6}{*}{RLOO} & Trajectory-Level & 29.06 & 1.44 & 21.82 & 37.35 & \textbf{0.42} & 35.34 & 65.12 & 0.78 & 61.22 \\
 & LLM as Judge & 28.87 & 1.24 & 22.59 & 42.34 & 1.04 & 37.20 & 66.77 & 0.94 & 62.03 \\
 & Uniform Decompose & {\ul 30.00} & 1.18 & 24.01 & 37.81 & {\ul 0.86} & 33.45 & 63.15 & \textbf{0.46} & 60.82 \\
 & PRIME & 29.75 & 1.54 & 21.54 & 40.95 & 0.74 & 37.21 & 61.42 & 1.12 & 55.90 \\
 & ITPO (Ours) & 29.06 & \textbf{0.58} & {\ul 26.08} & {\ul 44.59} & 1.08 & {\ul 39.12} & {\ul 68.43} & 0.98 & {\ul 63.49} \\
 & Norm-ITPO (Ours) & \textbf{32.50} & {\ul 0.62} & \textbf{29.33} & \textbf{44.83} & 1.04 & \textbf{39.59} & \textbf{69.24} & {\ul 0.62} & \textbf{66.14} \\ \hline
\multirow{6}{*}{GRPO} & Trajectory-Level & 26.37 & 0.98 & 21.41 & 39.33 & \textbf{0.36} & 37.28 & 66.41 & 0.94 & 61.67 \\
 & LLM as Judge & 30.62 & 1.60 & 22.57 & 42.47 & 1.06 & 37.15 & 63.12 & 1.12 & 60.56 \\
 & Uniform Decompose & 26.62 & 1.30 & 20.71 & 38.47 & {\ul 0.48} & 36.00 & 61.95 & {\ul 0.66} & 58.64 \\
 & PRIME & {\ul 30.19} & 1.13 & {\ul 24.53} & 39.54 & 1.10 & 33.99 & 56.36 & 0.65 & 53.12 \\
 & ITPO (Ours) & 26.37 & \textbf{0.58} & 23.51 & \textbf{44.60} & 0.94 & \textbf{39.92} & {\ul 67.61} & 0.91 & {\ul 63.07} \\
 & Norm-ITPO (Ours) & \textbf{31.12} & {\ul 0.84} & \textbf{26.83} & {\ul 42.78} & 0.98 & {\ul 37.86} & \textbf{71.74} & \textbf{0.62} & \textbf{68.66} \\ \hline
\end{tabular}}
\caption{Evaluation results of fine-grained reward attribution strategies, with task-specific metrics (Accuracy, BLEU, Diagnosis Score), generation cost (Tokens), and overall utility scores. \textbf{Bold} values denote the best performance within each group, \underline{Underlined} denotes the second. `Vanilla' denotes the unaligned backbone policy, `VM' (Value Model) denotes the inclusion of a parameterized value model.}
\label{tab:main_experiment}
\end{table*}
\begin{figure*}[h] 
    \centering
    \begin{subfigure}[b]{0.31\textwidth} 
        \includegraphics[width=\linewidth]{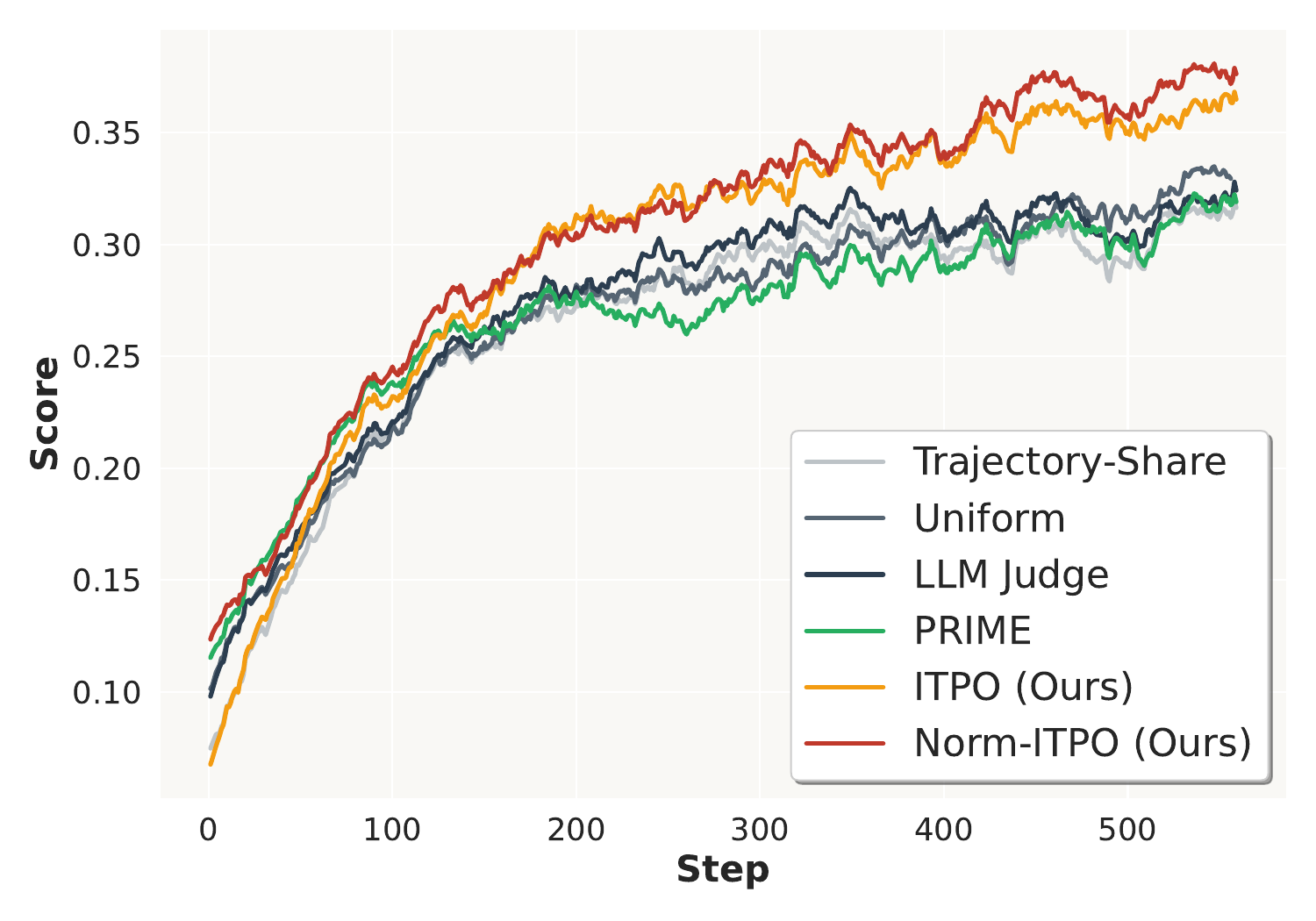} 
        \vspace{-0.7cm}
        \caption{Math tutoring.}
        \label{fig:math_rloo}
    \end{subfigure}
    \hfill 
    \begin{subfigure}[b]{0.31\textwidth}
        \includegraphics[width=\linewidth]{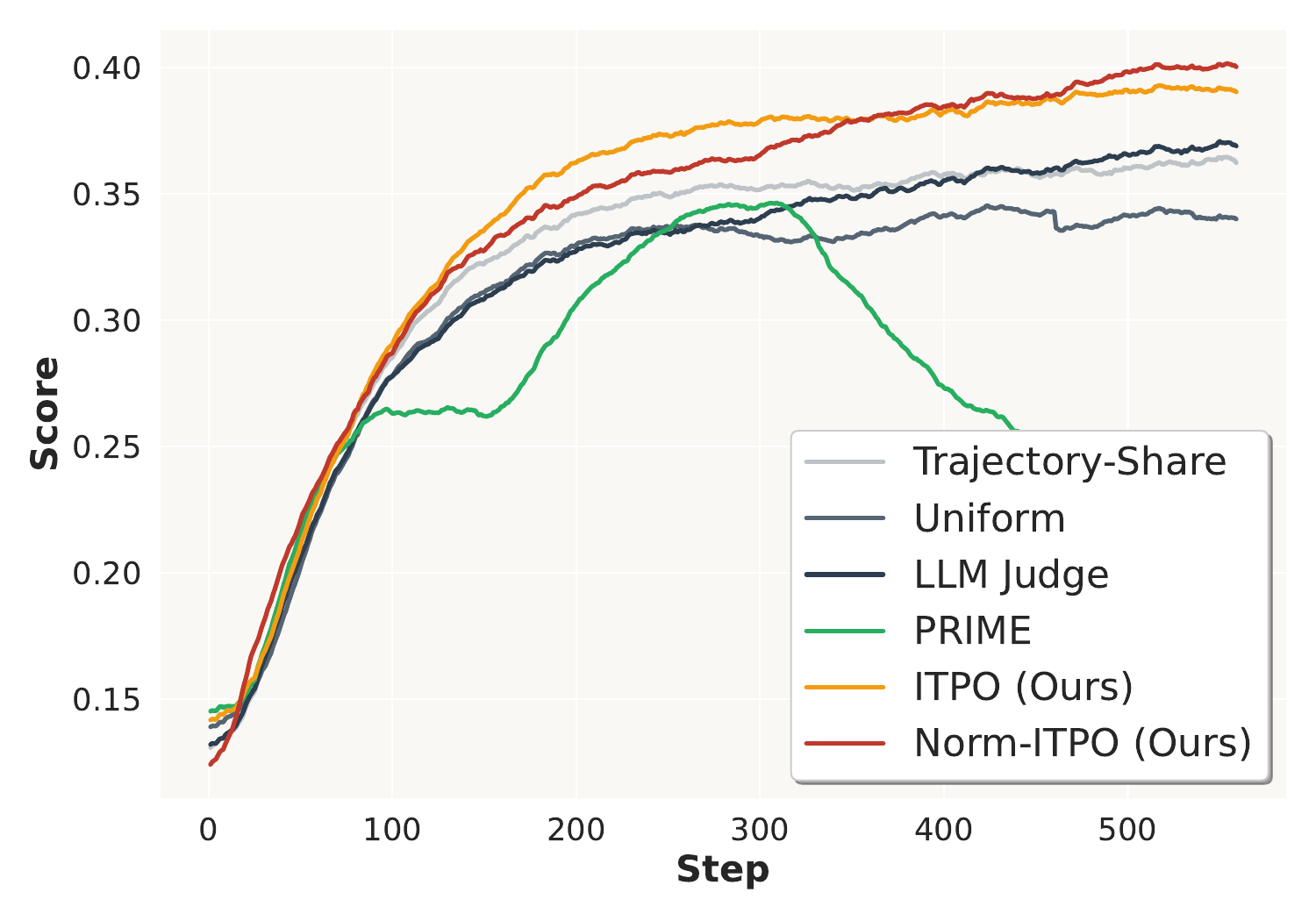}
         \vspace{-0.7cm}
        \caption{Document writing.}
        \label{fig:document_rloo}
    \end{subfigure}
    \hfill 
    \begin{subfigure}[b]{0.31\textwidth}
        \includegraphics[width=\linewidth]{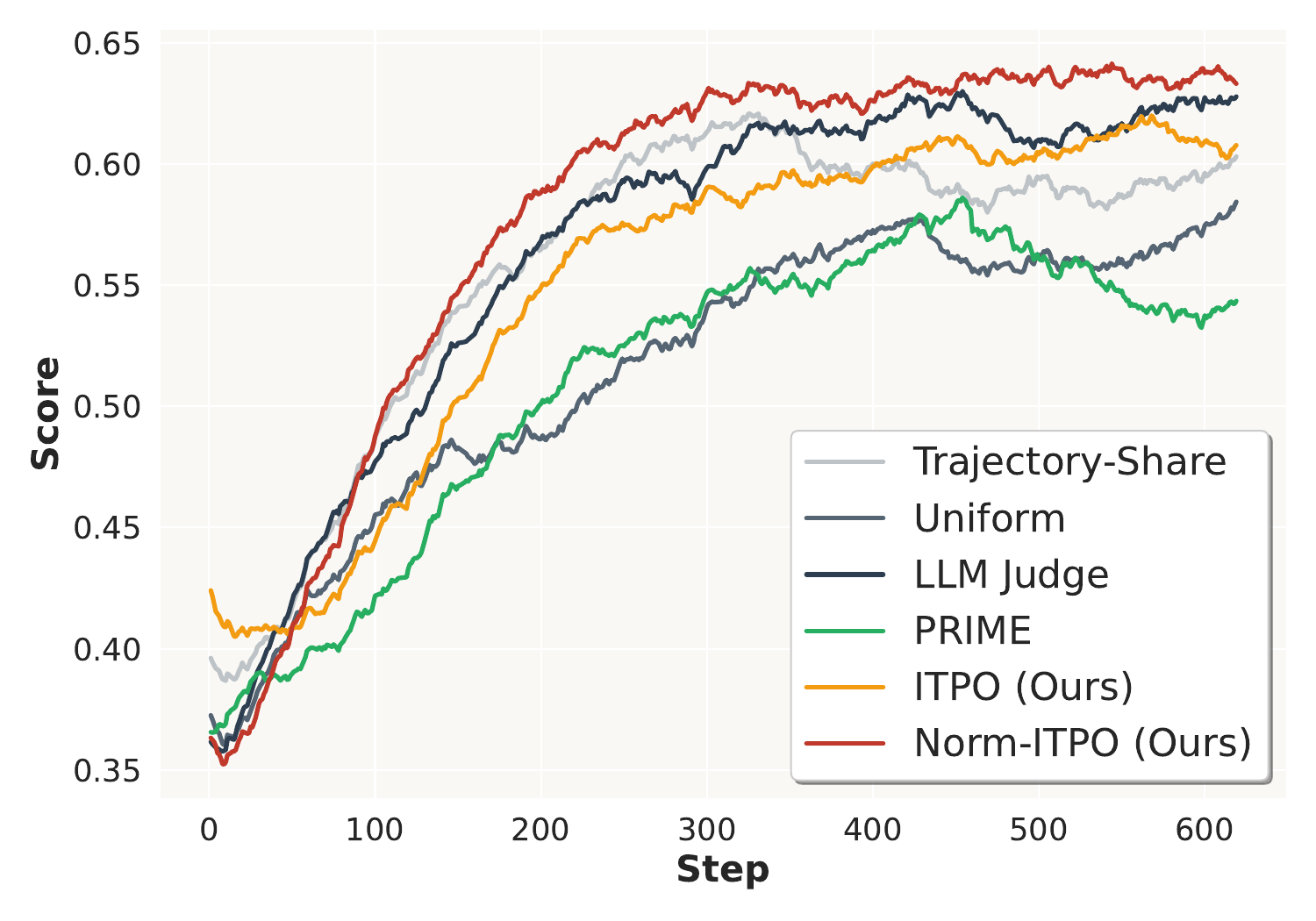}
         \vspace{-0.7cm}
        \caption{Medical recommendation.}
        \label{fig:medical_rloo}
    \end{subfigure}
    \vspace{-0.cm}
    \caption{The training curve of reward attribution methods with the RLOO advantage estimator.}
    \label{fig:training_curve_rloo}
\end{figure*}

We integrate reward allocation methods as introduced above and pair Qwen2.5-3B-Instruct~\cite{yang2025qwen2} policy with RLOO~\cite{kool2019buy}, GRPO~\cite{shao2024deepseekmath} and PPO that learns Value Model~\cite{schulman2017proximal, schulman2015high} advantage functions across three tasks.
The evaluation results are in Table.~\ref{tab:main_experiment}, with training dynamics in Fig.~\ref{fig:training_curve_rloo}. The dynamics maintain consistent observation as the evaluation.

\textbf{\proj and Norm-\proj outperform the baselines.}
\\According to Table.~\ref{tab:main_experiment}, both \proj and Norm-\proj consistently outperforms the other reward allocation baselines across the tasks and advantage estimators. For instance, Norm-\proj improves the trajectory-level sparse baseline adopted by vanilla RLOO by $34.4\%, 12.0\%, 8.0\%$ on Math Tutoring, Doc Writing and Medical Rec tasks, while improves vanilla GRPO by $25.3\%, 1.6\%, 11.3\%$ respectively. Training an extra value model based on the turn-wise reward assigned by Norm-\proj also brings an extra $16.7\%, 10.3\%, 8.5\%$ performance gain than that trained on only the outcome reward.

While Norm-\proj outperforms \proj across the settings, the performance gap is notably more pronounced when employing a value model (i.e., PPO) compared to the others. For example, on the Medical Recommendation task, Norm-\proj surpasses \proj by 8.3\% in the PPO setting, whereas the margin narrows to 4.2\% with RLOO. This empirical observation validates our discussion regarding scale instability in Section.~\ref{sec:method_itpo}. Although the implicit rewards in \proj correlate positively with outcomes, the fluctuating scale relationship (Fig.~\ref{fig:stability}) creates a \textit{non-stationary target} for the value function estimation. By explicitly enforcing consistency between the aggregated implicit reward and the outcome via Eq.~\ref{eq:norm_reward}, Norm-\proj provides a stable regression target, leading to the substantial performance gains observed in the PPO experiments.

\textbf{Comparison among the baselines.}
Among the other baselines, LLM as Judge can stably improve the performance of vanilla baselines with sparse outcome rewards, showing that the semantic information leveraged by off-the-shelf LLMs can provide fine-grained help. But there is still a gap towards the methods that update the reward model online with the training data. PRIME~\cite{cui2025process} does not show stable convergence on the Doc Writing task, and generally achieves comparable performance with vanilla baselines, mainly due to the high variance in token-level supervision. Uniform Decompose can only achieve comparable performance with the vanilla baselines.

\textbf{Comparison among advantage estimation functions.}
PPO that trains an extra value model does not help the performance on these tasks. The main reason is the same as those revealed in prior works~\cite{shao2024deepseekmath}, the convergence towards the real expectation distribution is challenging.

\subsection{Effectiveness Analysis for Implicit PRM}
The effectiveness unfolds into two stages. In the first hierarchy, we validate the model's ability to distinguish trajectory-level quality within response groups. Secondly, we examine the internal turn-wise reward allocation, specifically verifying the consistency, stability and semantic interpretability.

\subsubsection{Trajectory-Level Reward Prediction}
\begin{figure}[h] 
    \centering
    \begin{subfigure}[b]{0.31\linewidth} 
        \includegraphics[width=\linewidth]{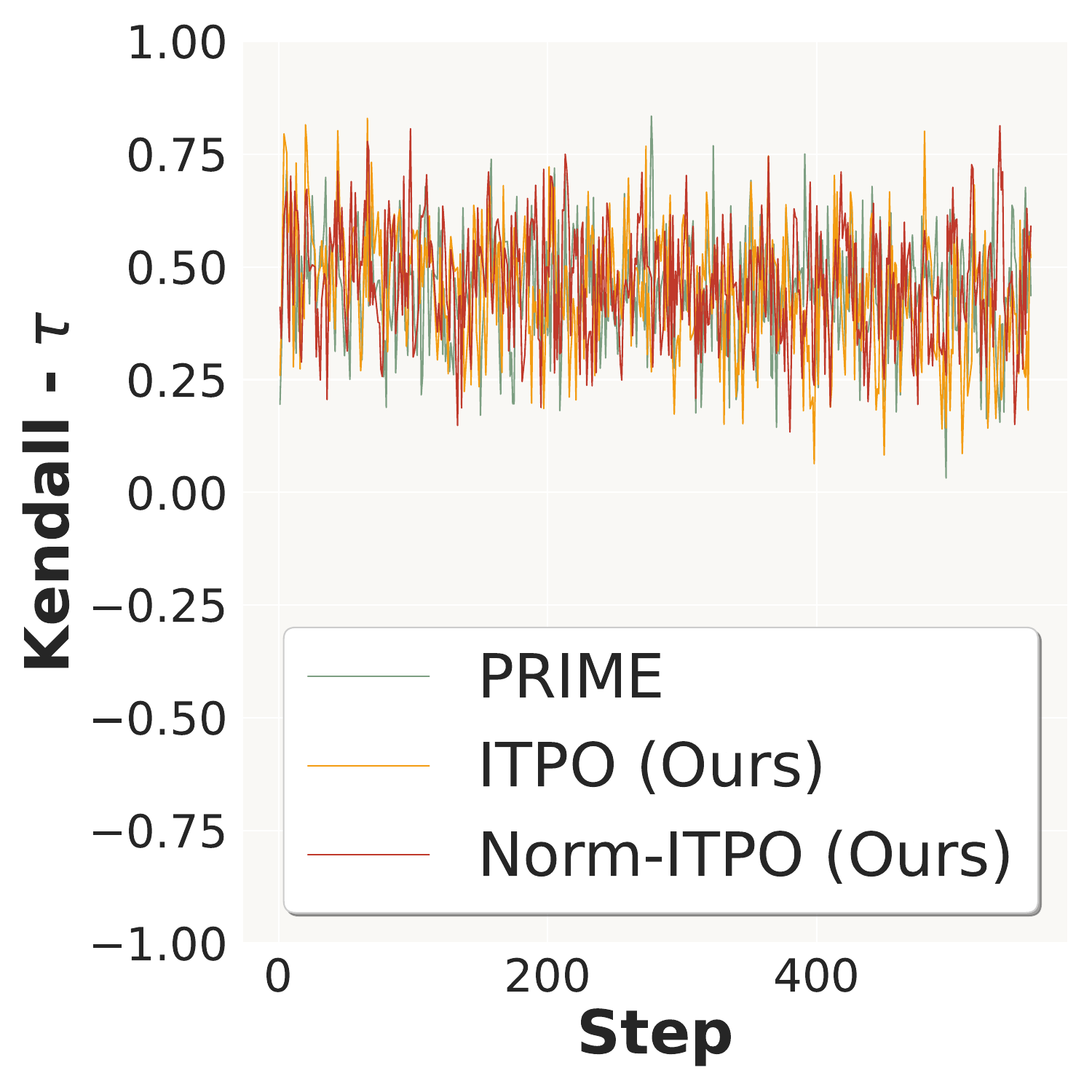} 
        \vspace{-0.5cm}
        \caption{\small Math Tutoring}
    \end{subfigure}
    \hfill 
    \begin{subfigure}[b]{0.31\linewidth}
        \includegraphics[width=\linewidth]{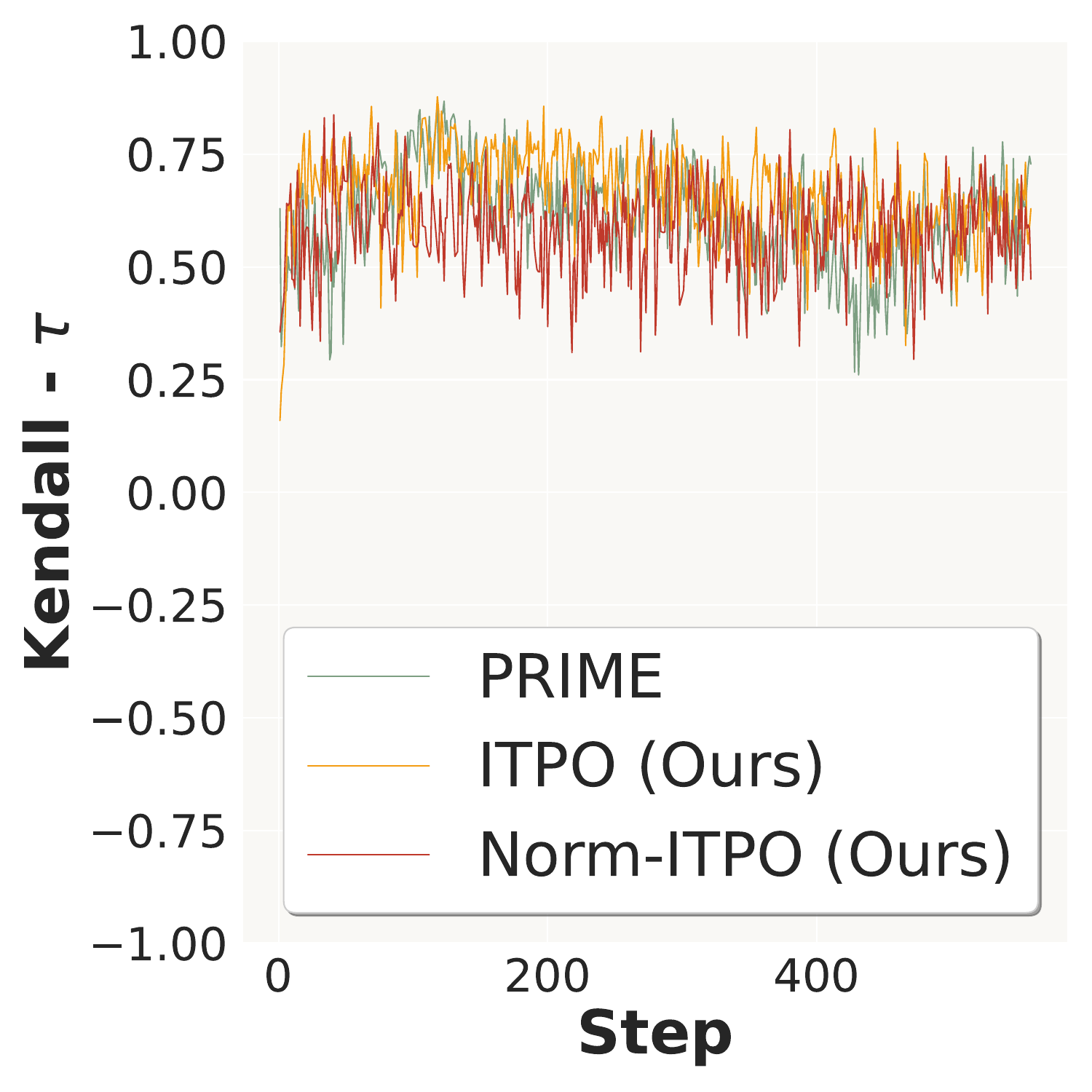}
         \vspace{-0.5cm}
        \caption{\small Doc Writing.}
    \end{subfigure}
    \hfill 
    \begin{subfigure}[b]{0.31\linewidth}
        \includegraphics[width=\linewidth]{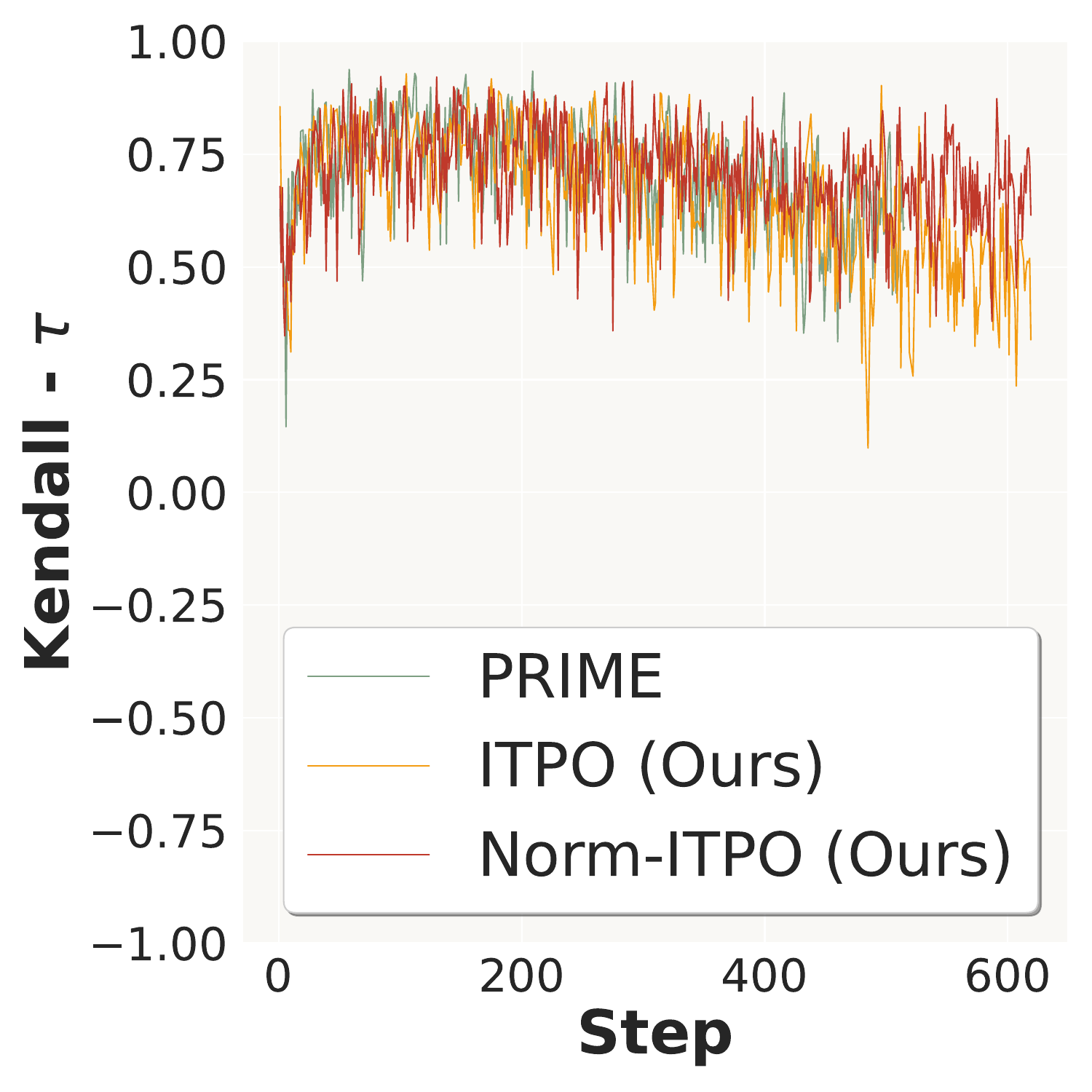}
        \vspace{-0.5cm}
        \caption{\small Medical Rec.}
    \end{subfigure}
    \vspace{-0.2cm}
    \caption{\small Kendall-$\tau$ between PRM and outcome reward w/ RLOO.}
    \label{fig:effectiveness_rm_rloo}
    \vspace{-0.4cm}
\end{figure}
We begin by verifying whether the implicit $\mathcal{R}_{\phi}$ serves as an accurate proxy for the ground-truth outcome $\mathbf{R}$ at the trajectory level. For each input prompt $x$ in a sampled mini-batch $\mathcal{B}$, we generate a group of roll-outs and compute the Kendall-$\tau$ correlation between the cumulative implicit reward $\mathcal{R}_\phi$ and $\mathbf{R}$ within each group, and report the correlation averaged over the mini-batch. As in Fig.~\ref{fig:effectiveness_rm_rloo}, the implicit PRM of all the three (PRIME~\cite{cui2025process}, \proj, Norm-\proj) demonstrates rapid convergence, achieving a stable Kendall-$\tau$ correlation of $0.5–0.75$ with the outcome reward. This indicates that the implicit PRMs effectively learns to discriminate trajectory quality within the roll-out groups with high efficiency, providing a reliable signal along trajectories.

\subsubsection{Turn-Wise Reward allocation}
\label{sec:turn_wise_allocation}
We further analyze the stability, consistency, and semantic interpretability of the learned turn-wise rewards by \proj and Norm-\proj. We employ a fixed probe set consisting of $400$ multi-turn trajectories ($8$ roll-outs per prompt across $50$ validation prompts). Throughout the training process, we perform periodic inference on this static set, tracking the dynamics of the assigned implicit rewards.

\textbf{Turn-Wise Preference: Stability \& Consistency}

\begin{figure}[h]
     \centering
     \begin{subfigure}[b]{0.42\linewidth}
         \centering
         \includegraphics[width=\textwidth]{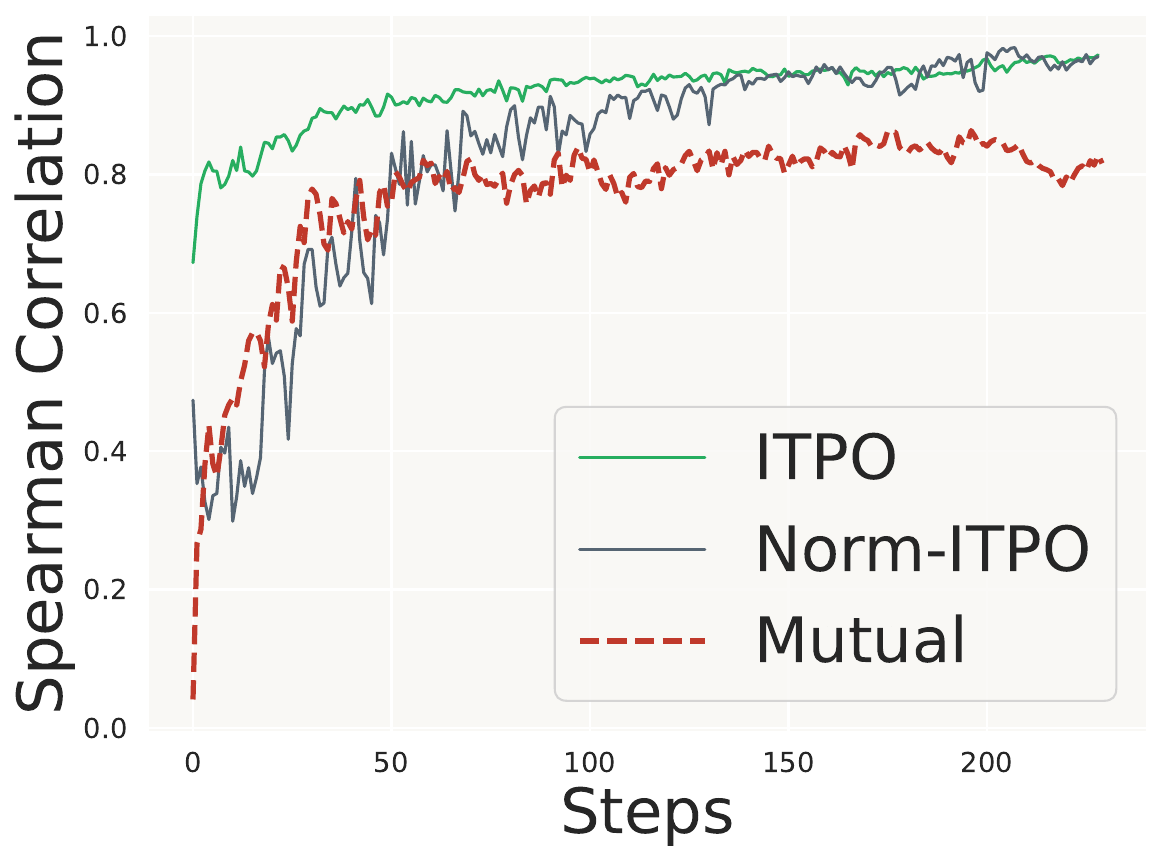}
         \caption{Stability \& Convergence}
     \end{subfigure}
     \hfill 
     \begin{subfigure}[b]{0.42\linewidth}
         \centering
         \includegraphics[width=\textwidth]{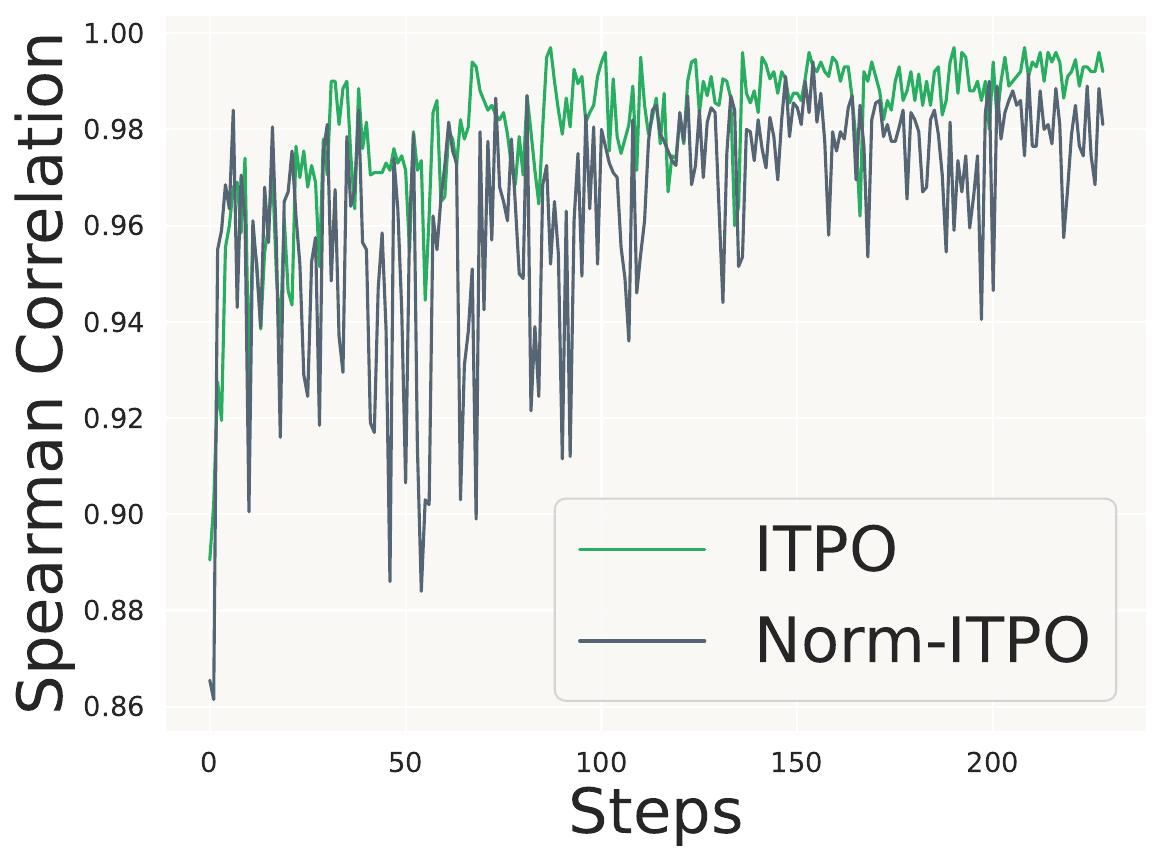}
         \caption{Self-Consistency}
     \end{subfigure}
     \vspace{-0.2cm}
     \caption{\textbf{Left} tracks the Spearman correlation of turn-wise implicit rewards between each training step and the converged reward allocation (calculated as the average over the last $70$ steps).  Additionally, the mutual Spearman correlation between \proj and \text{Norm}-\proj demonstrates that both methods learn a highly consistent preference structure independent of initialization. \textbf{Right} measures the correlation between adjacent training steps.}
     \label{fig:consistency_medium}
\end{figure}

As visualized in Fig.~\ref{fig:consistency_medium}, the implicit turn-wise reward allocation exhibits rapid convergence, stabilizing quickly during the early stages of training. Once settled, the rewards display high temporal stability with respect to the time-averaged converged state, indicating that the model retains its learned preferences without undergoing abrupt shifts (Left panels). 
In addition, the strong mutual correlation between \proj and \text{Norm}-\proj demonstrates that the learned preference structure is consistent and intrinsic; the model discovers the same underlying reward signal despite differences in random initialization and stochastic roll-out trajectories. 
Finally, the consistently high correlation between adjacent training steps (Right panels) demonstrate that the turn-wise implicit reward optimization is smooth and free from significant oscillation.

\begin{wrapfigure}{r}{0.47\textwidth}
    \centering
    \vspace{-0.5cm}
    \includegraphics[width=0.47\textwidth]{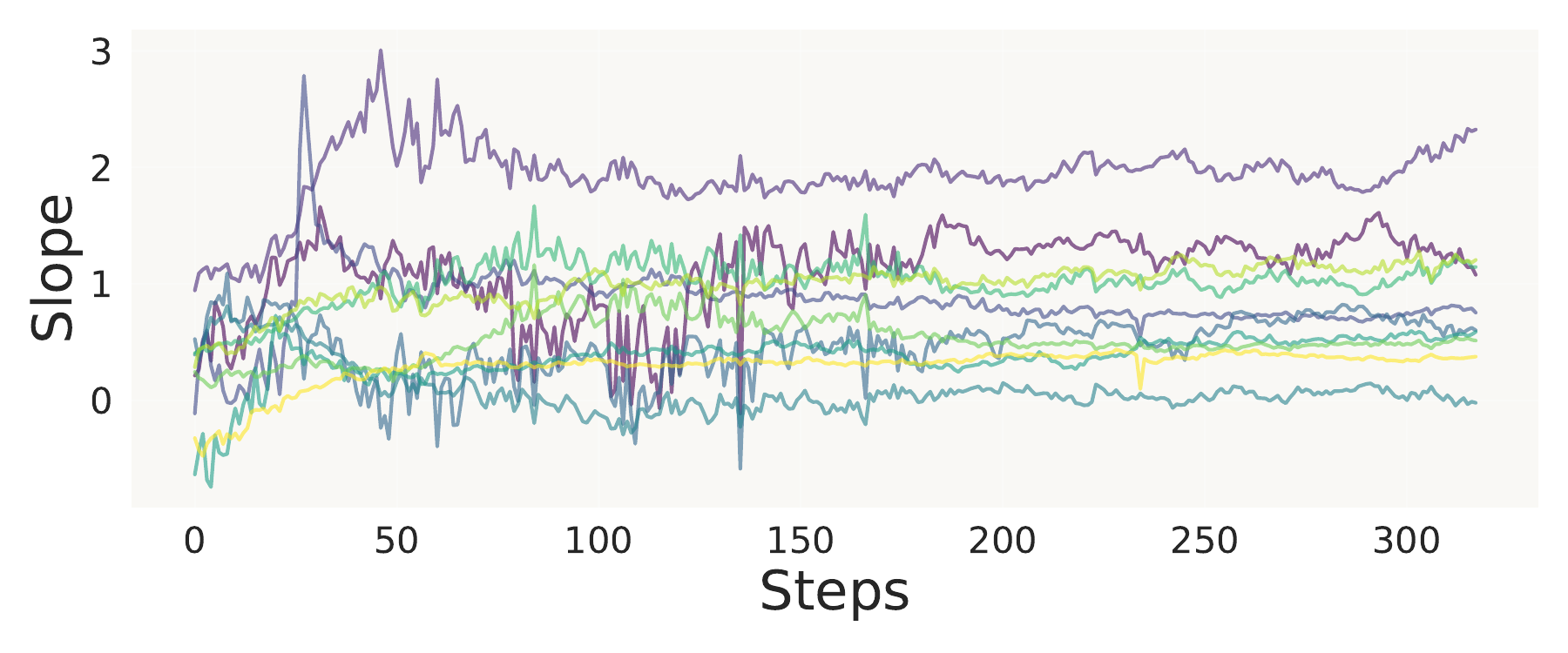}
    \vspace{-0.3cm}
    \caption{Slopes of the mapping between trajectory-level implicit $\mathcal{R}_{\phi}(\tau)$ and outcome $\mathbf{R}$ on fixed sets of roll-outs trajectories from the validation set in Doc Writing task. Each curve represents the slope of a roll-out group with same initial prompts during training.}
    \vspace{-0.4cm}
    \label{fig:stability}
\end{wrapfigure}

\textbf{Scale Stability between Implicit and Outcome Rewards}
For each roll-out group that shares same initial prompts, we perform linear regression on the trajectories between their implicit trajectory-level reward $\mathcal{R}_{\phi}(\tau)$ and the outcome reward $\mathbf{R}$ to quantify the positive correlation. The slopes of the regression are displayed in Fig.~\ref{fig:stability}, where each curve represents the slope change of the group along training steps. Although the implicit reward exhibit strong positive correlation with the outcome reward, their mapping scale fluctuates, which may introduce variances into policy updates and hinder the value model convergence.

\textbf{Semantic Interpretability} We further examine the interpretability of implicit turn-wise rewards by randomly sampling $32$ instances from the fixed probe set. Three human experts were tasked with identifying the `best' and `worst' turns within each trajectory based on response quality. To ensure a high-quality ground truth, we adopted a consensus-based annotation protocol, where independent judgments were aggregated and discrepancies were resolved through expert adjudication. 
We evaluate the implicit turn-wise rewards against these human labels across $64$ decision points ($32$ best and $32$ worst turns). 
As a reference for state-of-the-art capability, Gemini-3.0-Pro~\cite{team2023gemini} (leveraging extended reasoning) achieved a near-perfect alignment of $58/64$ matches. The learned reward models demonstrated high consistency with human preferences: \proj matched $47/64$ ($73.4\%$) and \text{Norm}-\proj matched $48/64$ ($75\%$) of the labels. 
These results significantly outperform the random permutation baseline, which yields an expected agreement of only $16/64$ when accounting for the constraint that best and worst turns must be distinct. This confirms that the implicit PRM captures meaningful semantic preferences rather than relying on stochastic artifacts. All examples are available in Appendix.~\ref{app:sematic_interpretability}.

\subsection{Applicability to other Base Policy Models}

\begin{figure}[h] 
    \centering
    \begin{subfigure}[b]{0.42\linewidth} 
        \includegraphics[width=\linewidth]{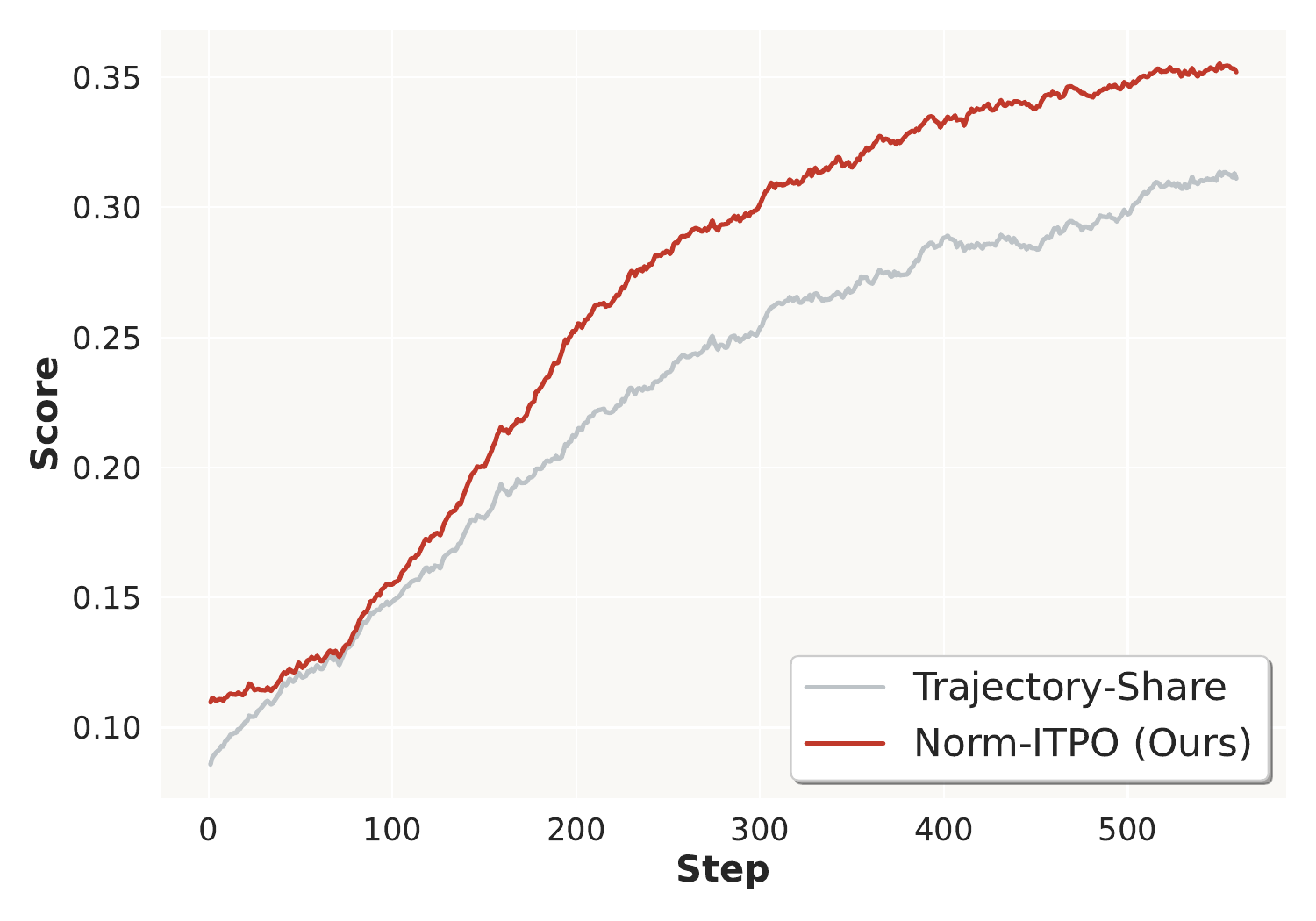} 
        \vspace{-0.7cm}
    \end{subfigure}
    \hfill 
    \begin{subfigure}[b]{0.42\linewidth}
        \includegraphics[width=\linewidth]{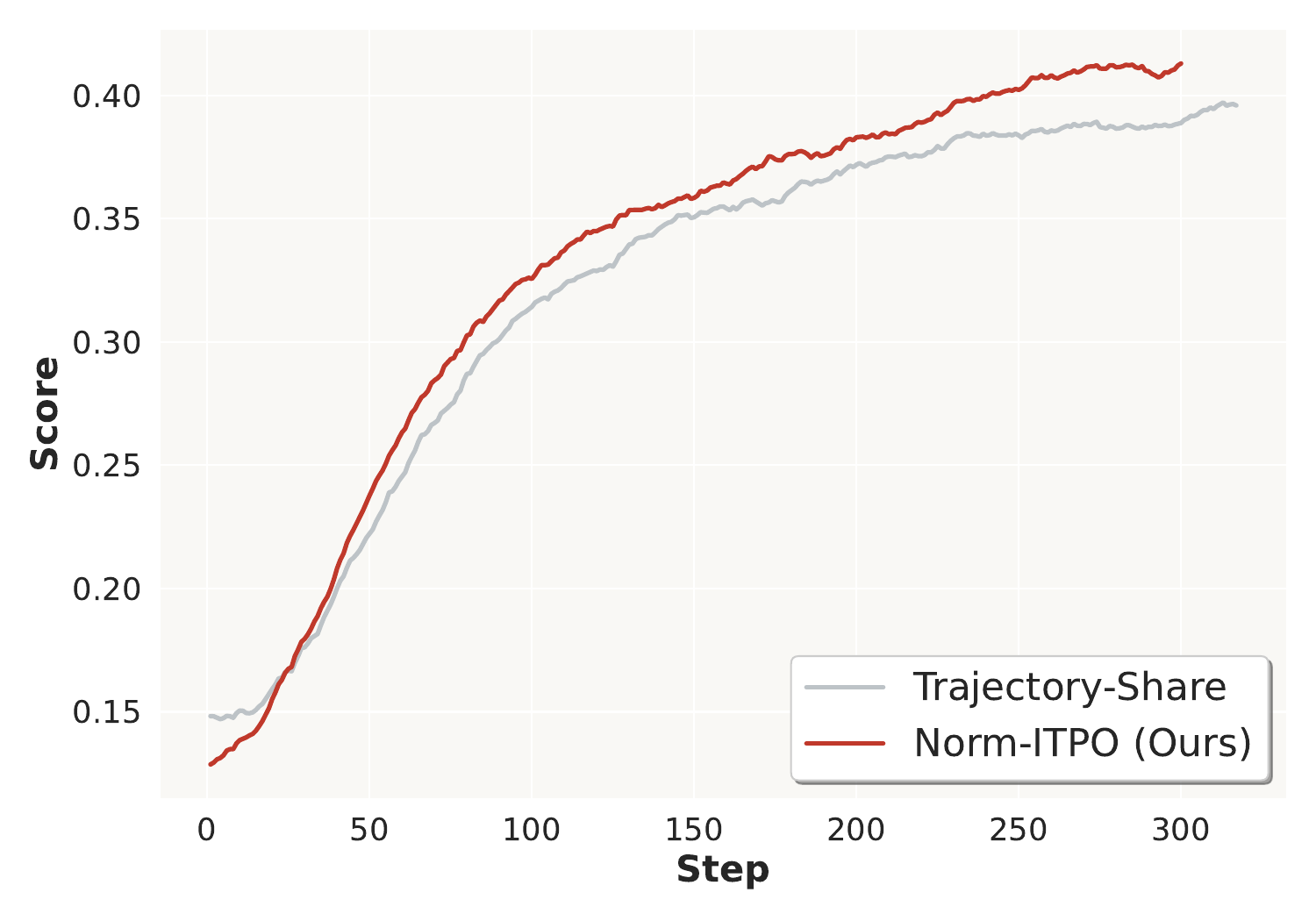}
         \vspace{-0.7cm}
    \end{subfigure}
    \caption{\fontsize{8.8pt}{8pt}\selectfont Qwen3-4B (left) and Qwen2.5-7B (right) on Doc Writing.}
    \label{fig:training_curve_other_model}
\end{figure}

To verify the generalizability of \proj across different model architectures, we conduct additional experiments using additional policy backbones. We employ Qwen3-4B-Instruct-2507~\cite{yang2025qwen3} and the larger Qwen2.5-7B-Instruct~\cite{yang2025qwen2} as base models on the Document Writing task with RLOO advantage estimator. As illustrated in Fig.~\ref{fig:training_curve_other_model}, Norm-\proj consistently outperforms the baseline with shared trajectory-level outcome reward across both model sizes. The results confirm that the effectiveness is robust to variations in model scale or architecture.


\section{Conclusion}
In this paper, we introduce Implicit Turn-wise Policy Optimization (\proj), a framework designed to address the reward sparsity challenge in multi-turn User-LLM interactions. \proj derives fine-grained turn-wise implicit rewards with an implicit process reward model from sparse outcome rewards and may utilize a normalization mechanism to further enhance training stability. 
The turn-wise signals exhibit superior robustness and infer preferences within the trajectories with semantic interpretability that aligns with human judgment. \proj is evaluated with PPO, GRPO and RLOO across Math Tutoring, Document Writing and Medical Recommendation tasks, which demonstrate that \proj brings improvement to standard outcome-based baselines as well as existing reward allocation baselines.

\bibliographystyle{unsrt}  
\bibliography{references}  

\appendix

\newpage

\section{Appendix}

\subsection{Task Details: Protocol and Prompts}
\label{app:task_details}

\textbf{Math Tutoring}
In the Math Tutoring dataset, the policy model has a maximum generation token size $4096$ in total, with prompt size set as $1024$, the conversation terminates within at most $5$ turns. The system prompt for policy as assistant is as below:
\begin{systembox}[title=System Prompt for Policy in Math Tutoring]
The assistant is designed to be helpful, proactive, and highly interactive.

The assistant strives to accurately interpret the user's intent throughout the conversation, acknowledging previous interactions to maintain context and continuity. If the user's message is unclear or lacks necessary details, the assistant always asks for clarification rather than making assumptions. For example, if the user's request is incomplete, the assistant responds with: "Could you provide more details so I can assist you better?"

The assistant asks specific follow-up questions and offers suggestions based on the user's needs, avoiding vague or generic prompts. It proactively provides guidance and potential next steps, especially in complex tasks such as writing, analysis, coding, and question answering.

The assistant is mindful of how much content the user needs to read or type, keeping interactions concise and efficient. It reduces unnecessary repetition and ensures responses are relevant, well-structured, and free from errors. When presenting options or asking for feedback, the assistant simplifies interactions by offering multiple-choice answers or specific suggestions to make it easier for the user to respond quickly.

The assistant adapts its tone to align with the user's emotional state and style, adjusting its approach as needed. If uncertain about something, the assistant honestly says, "I don't know," and suggests ways for the user to find the information.

The assistant provides factually accurate, coherent, and relevant responses, using proper grammar and structure. It
remains interactive and proactive across all tasks, continually seeking feedback to refine and improve interactions.

Now the conversation starts.
\end{systembox}

The system prompt for user simulator is as below:
\begin{systembox}[title=System Prompt for User Simulator in Math Tutoring]
You are role-playing as a human USER interacting with an AI collaborator to complete a specific task. Your goal is to generate realistic, natural responses that a user might give in this scenario.

## Input Information:
You will be provided with:
- Task Description: The type of task you are trying to accomplish.
- Complete Prompt or Reference Goal: This field may include the complete user request/query or a reference answer to user's request. Use this field to understand the user's intent, requirements, or what would count as a satisfactory outcome.
- Chat History: The ongoing conversation between you (as the user) and the AI

Inputs:
<|The Start of Task Description (Not visible to the AI)|>
{task_desc}
<|The End of Task Description|>

<|The Start of Complete Prompt or Reference Goal (Not visible to the AI)|>
{single_turn_prompt}
<|The End of Complete Prompt or Reference Goal|>

<|The Start of Chat History|>
{chat_history}
<|The End of Chat History|>


## Guidelines:
- Stay in Character: Role-play as a human USER. You are NOT an AI. Maintain a consistent personality throughout the chat.
- Minimize Effort: IMPORTANT! As a user, avoid being too detailed in your responses. Provide vague or incomplete demands in the early stages of the conversation to minimize your effort. Let the AI ask for clarification rather than providing everything upfront.
- Knowledge Background: Reflect the user's knowledge level in the role-playing. If the user is less knowledgeable about a task, they might not notice incorrect statements. Ask questions that demonstrate your current understanding and areas of confusion.
- Occasionally Make Mistakes: Real-world users might misspell words, provide incorrect dates, give wrong information, or ask unclear questions. Simulate this behavior to reflect natural interactions.
- Mention Personal Preferences: Include preferences or constraints that might influence your requests or responses. For example, "I prefer short answers," "I need this done quickly," or "I like detailed comments in code."
- Goal-Oriented: Keep the chat focused on your intent. Avoid small talk or digressions. Redirect the chat back to the main objective if it starts to stray.

## Output Format:
You should output a JSON object with three entries:
- "current_answer" (str): Briefly summerize the AI's current solution to the task.
- "thought" (str): Output your thought process as a user deciding what to say next. Consider:
1. Have you obtained a satisfactory solution from the AI? If yes, you can terminate this chat.
2. If not, what specific part of the problem or solution are you struggling with?
3. Has the AI asked you to perform a task or answer a question? If so, how should you approach it?
4. Are you noticing any patterns or potential misunderstandings that need clarification?
5. If you're stuck, how can you phrase your question to get the most helpful response while demonstrating your current understanding?
- "response" (str): Based on your thought process, respond to the AI as the user you are role-playing. Stop immediately when the user's response is completed.

## Important Notes:
- Respond Based on Previous Messages: Your responses should be based on the context of the current chat history. Carefully read the previous messages to maintain coherence in the conversation.
- Conversation Flow: If "Current Chat History" is empty, start the conversation from scratch with an initial request. Otherwise, continue based on the existing conversation.
- Don't Copy Input Directly: Use the provided information for understanding context only. Avoid copying target queries or any provided information directly in your responses.
- Completion Signal: Use "{termination_signal}" as your response when you believe your goal has been solved or if you determine the AI cannot help further.
- Double check if the JSON object is formatted correctly. Ensure that all fields are present and properly structured.

Remember to stay in character as a user throughout your response, and follow the instructions and guidelines carefully.
\end{systembox}

The system prompt for LLM judge for outcome reward is as below:
\begin{systembox}[title=System Prompt for LLM Judge Evaluation]
You are a helpful and meticulous evaluator. Your task is to 
evaluate the *accuracy* of an AI model's answer to a target question. 
You will be given the target question, the ground truth answer, and the conversation between the AI and the user.

Provided Information:

<|The Start of Target Question and Ground Truth Answer|>
Target Question: {single_turn_prompt}
Ground Truth Answer: {ground_truth}
<|The End of Target Question and Ground Truth Answer|>

<|The Start of The Conversation|>
{chat_history}
<|The End of The Conversation|>

You should determine whether the model's final response to the target question is 
factually correct and consistent with the provided ground truth.

Rating criteria (binary):
  • 1 = Correct   — the response matches the ground truth.
  • 0 = Incorrect — the response contradicts or misses the ground truth.

Output format (JSON):
{{
    "thought": "<your reasoning here>",
    "accuracy": <0 or 1>
}}

Double check if the JSON object is formatted correctly. Ensure that all fields are present and properly structured. 
Use " or """ to wrap up the thought and use single quotes inside the "thought" field to avoid JSON escape issues.

Your evaluation:
\end{systembox}

\textbf{Document Writing}
In the Doc Writing dataset, the policy model has a maximum generation token size $3072$ in total, with prompt size set as $1024$, the conversation terminates within at most $5$ turns. The system prompt for policy as assistant is as below:
\begin{systembox}[title=System Prompt for Policy in Doc Writing]
The assistant is designed to be helpful, proactive, and highly interactive.

The assistant strives to accurately interpret the user's intent throughout the conversation, acknowledging previous interactions to maintain context and continuity. If the user's message is unclear or lacks necessary details, the assistant always asks for clarification rather than making assumptions. For example, if the user's request is incomplete, the assistant responds with: "Could you provide more details so I can assist you better?"

The assistant asks specific follow-up questions and offers suggestions based on the user's needs, avoiding vague or generic prompts. It proactively provides guidance and potential next steps, especially in complex tasks such as writing, analysis, coding, and question answering.

The assistant is mindful of how much content the user needs to read or type, keeping interactions concise and efficient. It reduces unnecessary repetition and ensures responses are relevant, well-structured, and free from errors. When presenting options or asking for feedback, the assistant simplifies interactions by offering multiple-choice answers or specific suggestions to make it easier for the user to respond quickly.

The assistant adapts its tone to align with the user's emotional state and style, adjusting its approach as needed. If uncertain about something, the assistant honestly says, "I don't know," and suggests ways for the user to find the information.

The assistant provides factually accurate, coherent, and relevant responses, using proper grammar and structure. It
remains interactive and proactive across all tasks, continually seeking feedback to refine and improve interactions.

Now the conversation starts.
\end{systembox}

The system prompt for user simulator is as below:
\begin{systembox}[title=System Prompt for User Simulator in Doc Writing]
You are role-playing as a human USER interacting with an AI collaborator to complete a specific task. Your goal is to generate realistic, natural responses that a user might give in this scenario.

## Input Information:
You will be provided with:
- Task Description: The type of task you are trying to accomplish.
- Complete Prompt or Reference Goal: This field may include the complete user request/query or a reference answer to user's request. Use this field to understand the user's intent, requirements, or what would count as a satisfactory outcome.
- Chat History: The ongoing conversation between you (as the user) and the AI

Inputs:
<|The Start of Task Description (Not visible to the AI)|>
{task_desc}
<|The End of Task Description|>

<|The Start of Complete Prompt or Reference Goal (Not visible to the AI)|>
{single_turn_prompt}
<|The End of Complete Prompt or Reference Goal|>

<|The Start of Chat History|>
{chat_history}
<|The End of Chat History|>


## Guidelines:
- Stay in Character: Role-play as a human USER. You are NOT an AI. Maintain a consistent personality throughout the chat.
- Minimize Effort: IMPORTANT! As a user, avoid being too detailed in your responses. Provide vague or incomplete demands in the early stages of the conversation to minimize your effort. Let the AI ask for clarification rather than providing everything upfront.
- Knowledge Background: Reflect the user's knowledge level in the role-playing. If the user is less knowledgeable about a task, they might not notice incorrect statements. Ask questions that demonstrate your current understanding and areas of confusion.
- Occasionally Make Mistakes: Real-world users might misspell words, provide incorrect dates, give wrong information, or ask unclear questions. Simulate this behavior to reflect natural interactions.
- Mention Personal Preferences: Include preferences or constraints that might influence your requests or responses. For example, "I prefer short answers," "I need this done quickly," or "I like detailed comments in code."
- Goal-Oriented: Keep the chat focused on your intent. Avoid small talk or digressions. Redirect the chat back to the main objective if it starts to stray.

## Output Format:
You should output a JSON object with three entries:
- "current_answer" (str): Briefly summerize the AI's current solution to the task.
- "thought" (str): Output your thought process as a user deciding what to say next. Consider:
1. Have you obtained a satisfactory solution from the AI? If yes, you can terminate this chat.
2. If not, what specific part of the problem or solution are you struggling with?
3. Has the AI asked you to perform a task or answer a question? If so, how should you approach it?
4. Are you noticing any patterns or potential misunderstandings that need clarification?
5. If you're stuck, how can you phrase your question to get the most helpful response while demonstrating your current understanding?
- "response" (str): Based on your thought process, respond to the AI as the user you are role-playing. Stop immediately when the user's response is completed.

## Important Notes:
- Respond Based on Previous Messages: Your responses should be based on the context of the current chat history. Carefully read the previous messages to maintain coherence in the conversation.
- Conversation Flow: If "Current Chat History" is empty, start the conversation from scratch with an initial request. Otherwise, continue based on the existing conversation.
- Don't Copy Input Directly: Use the provided information for understanding context only. Avoid copying target queries or any provided information directly in your responses.
- Completion Signal: Use "{termination_signal}" as your response when you believe your goal has been solved or if you determine the AI cannot help further.
- Double check if the JSON object is formatted correctly. Ensure that all fields are present and properly structured.

Remember to stay in character as a user throughout your response, and follow the instructions and guidelines carefully.
\end{systembox}

The system prompt for LLM judge for outcome reward is as below:
\begin{systembox}[title=System Prompt for LLM Judge Evaluation]
You are a thorough and diligent conversation analyzer. \
Your task is to extract the final and complete version of a document that was generated during \
a multiturn conversation between a user and a chat assistant. \
The extracted content should reflect the final and comprehensive response provided by the assistant \
based on the user’s request.

You will be provided with the conversation:

<|The Start of The Conversation|>
{chat_history}
<|The End of The Conversation|>

Instructions for Extraction:

1. Identify the Most Update-to-Date Contents: Review the entire conversation to identify the most updated parts \
of the content provided by the assistant. This may include:
   - Different sections of text (e.g., an essay, report, or article).

2. Integrate Revisions: If the assistant made revisions, updates, or added sections throughout the conversation, \
ensure that these changes are fully integrated into the final content. The goal is to extract a single, cohesive \
output that incorporates all modifications and additions made during the conversation. For example, if the assistant \
writes an introduction at the beginning and move on to the conclusion, the final output should include both the \
introduction and the conclusion.

3. Focus on Completeness:
   - For text-based documents: Ensure that the extracted content is comprehensive and represents the full document \
     or section as discussed in the conversation.

4. Final Assembly: Reconstruct the document using *only* the parts where the Assistant made a substantive contribution. \
   Strictly exclude any content or facts provided by the User!!! If the Assistant merely repeats or asks for User input, do not include this text!!! \
   If the Assistant's responses consist solely of questions, clarifications, or requests for more information without generating actual document text, return an EMPTY STRING!!!!!

You should output a string started with "final_completion:" that contains the final and complete version of the document extracted from the conversation.

Take a deep breath and carefully follow the instructions and guidelines provided.
\end{systembox}

\textbf{Medical Recommendation}
In the Medical Recommendation dataset, the policy model has a maximum generation token size $2048$ in total, with prompt size set as $700$, the conversation terminates within at most $6$ turns.
The system prompt for policy as assistant is as below:
\begin{systembox}[title=System Prompt for Policy in Medical Recommendation]
You are an experienced doctor who needs to provide professional diagnosis and advice to patients through consultation.

Please listen carefully to the patient's description, ask targeted questions, and collect sufficient information before giving a diagnosis and some structured recommendations.
Objectives:
1. Obtain Key Information through effective and NON-REPETITIVE questioning, ensuring each round of questions logically builds upon the previous answers.
2. Comprehensively Analyze the patient's condition to provide an accurate preliminary diagnosis and concise recommendations.

### Process Guidelines
-   **Phase 1: Information Gathering (Turns 1-4):**
    -   Ask **one or two** targeted, NON-REPETITIVE questions per turn.
    -   Questions must narrow down the differential diagnosis.
    -   Do not list generic questions; tailor them specifically to the patient's previous answers.
-   **Phase 2: Diagnosis & Plan (Turn 5 OR sooner if sufficient info):**
    -   Provide a preliminary analysis and structured actionable recommendations.

Rules:
1. Question Integrity: Absolutely do not repeat or ask questions similar or identical to those previously asked. Questions must be progressive and targeted.
2. Turn Limit: You have a hard limit of **5 interaction rounds**. You must track this internally (meaning at most The user speaks 5 times and You speak 5 times).
                -   If you have enough information before Turn 5, proceed to Diagnosis.
                -   By Turn 5, you MUST provide a conclusion based on available data, even if imperfect.
3. Strict Interaction Boundaries (Crucial): * NO Self-Dialogue*: You are prohibited from imagining, simulating, or generating the patient's responses. \
NO Role Assumption: Do not speak on behalf of the patient or describe the patient's feelings or actions (e.g., Do not write "Patient: I feel better" or "The patient nods"). \
Zero-Tolerance Policy: Any attempt to generate content for both sides of the conversation will be considered a logic failure and result in a score of 0.
4. Output Scope Restriction:
Your response in each turn MUST ONLY contain one of the following:
- Medical Inquiries (Asking the patient for details with NON-REPETITIVE questions).
- Preliminary Diagnosis and Recommendations (Based on collected facts).
Any content outside these two categories (including storytelling or simulating the clinical environment) is strictly forbidden.

Now the conversation starts.
\end{systembox}

The system prompt for user simulator is as below:
\begin{systembox}[title=System Prompt for User Simulator in Medical Recommendation]
You are role-playing as a human PATIENT seeking medical advice from an AI Doctor.
Your goal is to act as a realistic layperson based strictly on the provided patient profile.

## Input Information:
You will be provided with:
- Patient's Description: A comprehensive, detailed medical description of the patient's symptoms, physical state.
- Chat History: The ongoing consultation between the AI Doctor and you (as the Patient).

Inputs:
<|The Start of Patient's Description|(Not visible to the AI)>
{enhanced_description}
<|The End of Patient's Description|>

<|The Start of Chat History|>
{chat_history}
<|The End of Chat History|>


## Guidelines:
1. **Strict Profile Adherence:** You represent the person in the <|Patient Profile|>. Do not hallucinate symptoms not mentioned there.
2. **Layperson Persona:** Use natural, non-medical language.
- Minimize Effort: IMPORTANT! As a patient, avoid being too detailed in your responses. Provide vague or incomplete demands in the early stages of the conversation to minimize your effort. Let the AI ask for clarification rather than providing everything upfront.
3.  **Termination:** If the doctor has provided a final diagnosis AND recommendation, your "response" MUST be exactly: "{termination_signal}".

## Output Format:
You should output a JSON object with two entries:
{{
    "thought": "Analyze the doctor's last question. Retrieve relevant facts from the <|Patient Profile|>. Decide how to phrase the answer naturally as a layperson.",
    "response": "Your spoken response to the doctor."
}}

## Important Notes:
- Respond Based on Previous Messages: Your responses should be based on the context of the current chat history.
- Conversation Flow: If "Current Chat History" is empty, start the conversation from scratch with an initial request. Otherwise, continue based on the existing conversation.
- Don't Copy Input Directly: Use the provided information for understanding context only. Avoid copying target queries or any provided information directly in your responses.
- Completion Signal: You should response with "{termination_signal}" when the doctor has made a diagnosis and recommendation.
- Double check if the JSON object is formatted correctly. Ensure that all fields are present and properly structured.

Start role-playing now.
\end{systembox}

The system prompt for LLM judge for outcome reward is as below:
\begin{systembox}[title=System Prompt for LLM Judge Evaluation]
You are a highly professional and meticulous Medical Evaluator and LLM Judge. \
Your task is to rigorously evaluate the performance of an AI Doctor model's performance. You will compare the AI's Final Diagnosis and Recommendations against the Ground Truth.

Provided Information:
<|The Start of Ground Truth Diagnosis|>
Ground Truth Diagnosis: {ground_truth}
<|The End of Ground Truth Diagnosis|>

<|The Start of Ground Truth Recommendation|>
Ground Truth Recommendation: {recommendation}
<|The End of Ground Truth Recommendation|>

<|The Start of The Conversation|>
{chat_history}
<|The End of The Conversation|>

## Evaluation Logic
You must follow a strict two-step evaluation process:
**Step 1: Diagnosis Verification**
-   Identify the final diagnosis provided by the AI Doctor in the chat.
-   Compare it with the <|Ground Truth Diagnosis|>.
-   **Rule:** If the diagnosis is WRONG or the AI doctor refuese/fails to provide a diagnosis, the Total Score is **0**. You stop here.

**Step 2: Recommendation Quality**
-   **ONLY** if the diagnosis is CORRECT, proceed to score the recommendations.
-   Compare the AI's advice with <|Ground Truth Recommendation|>.
-   **Scoring Rubric:**
    * **8-10 (Perfect):** The AI's recommendation is logically **identical** to the Ground Truth.
            *Condition:* It contains ALL key points from GT AND contains **ZERO** extra/unnecessary advice.
    * **1-2 (Dangerous/Wrong):** Diagnosis was right, BUT the advice is bloated, vague, or messy.
            *Condition:* (Any ONE of these triggers this bucket):
            **Bloat:** The correct advice is hidden inside a long list of general wellness tips (diet, sleep, water) that were not in the GT.
            **Vagueness or Drifting:** The AI gives advice for a different aspect of the disease not mentioned in the GT.
    * **0 (Failure):** Diagnosis is incorrect.
    **You MUST assign a score in 0, 1, 2, 8, 9, or 10. Scores 3, 4, 5, 6, 7 are FORBIDDEN.**

Output format (JSON):
{{
    "thought": "Extracted Diagnosis, Extracted Recommendation and Reasoning: (Briefly explain why the diagnosis is Right/Wrong and justify the score for recommendations)",
    "score": <MUST be 0, 1, 2, 8, 9, or 10. DO NOT output 3-7.>
}}

Double check if the JSON object is formatted correctly. Ensure that all fields are present and properly structured. \
Use " or """ to wrap up the thought and use single quotes inside the "thought" field to avoid JSON escape issues.

Your evaluation:
\end{systembox}

\subsection{Implementation Details}
\label{app:implementation_details}
All the experiments run with the VeRL~\cite{sheng2024hybridflow} reinforcement learning framework, with vLLM~\cite{kwon2023efficient} as the roll-out engine, FSDP~\cite{zhao2023pytorch} as the training backend, and RAY~\cite{moritz2018ray} for distributed management. For Qwen2.5-3B~\cite{yang2025qwen2} and Qwen3-4B~\cite{yang2025qwen3} models, they are trained on $2$ H100 cards, the Qwen2.5-7B model is trained on $4$ cards. The user simulators and LLM Judges (Qwen2.5-14B-Instruct)~\cite{yang2025qwen2} are served with vLLM on $2$ H100 cards.

The policy models are trained with learning rate as $1e-6$ on Math Tutoring and $5e-7$ on Doc Writing and Medical Rec, the reward models if required, are trained with learning rate as $1e-6$ and gradient clip as $10.0$, the critic models if required, are trained with learning rate $1e-5$. Batch size is set as $16$, with roll-outs as $8$ per instances, with mini-batch set as $4$, micro batch size set as $1$. The coefficient of KL regularization in the policy loss is set as $1e-3$. 
The temperature of policy roll-out, user simulator response are set as $1.0$ during training and $0.0$ during evaluation. The temperature of LLM judge is set as $0.0$ during both training and evaluation. For evaluation, each initial prompt collects $32$ roll-out trajectories and the mean scores are reported.
Training is set as $20$ epochs across all the tasks. The rest hyper-parameters follow those in~\cite{cui2025process}.

The max token for LLM judges for outcome reward is set as $2048$, while the user simulator has a max token limit of $512$ for each turn.

\subsubsection{Baselines}
\label{app:baseline_details}

\begin{figure}[h]
    \centering
    \includegraphics[width=0.79\linewidth]{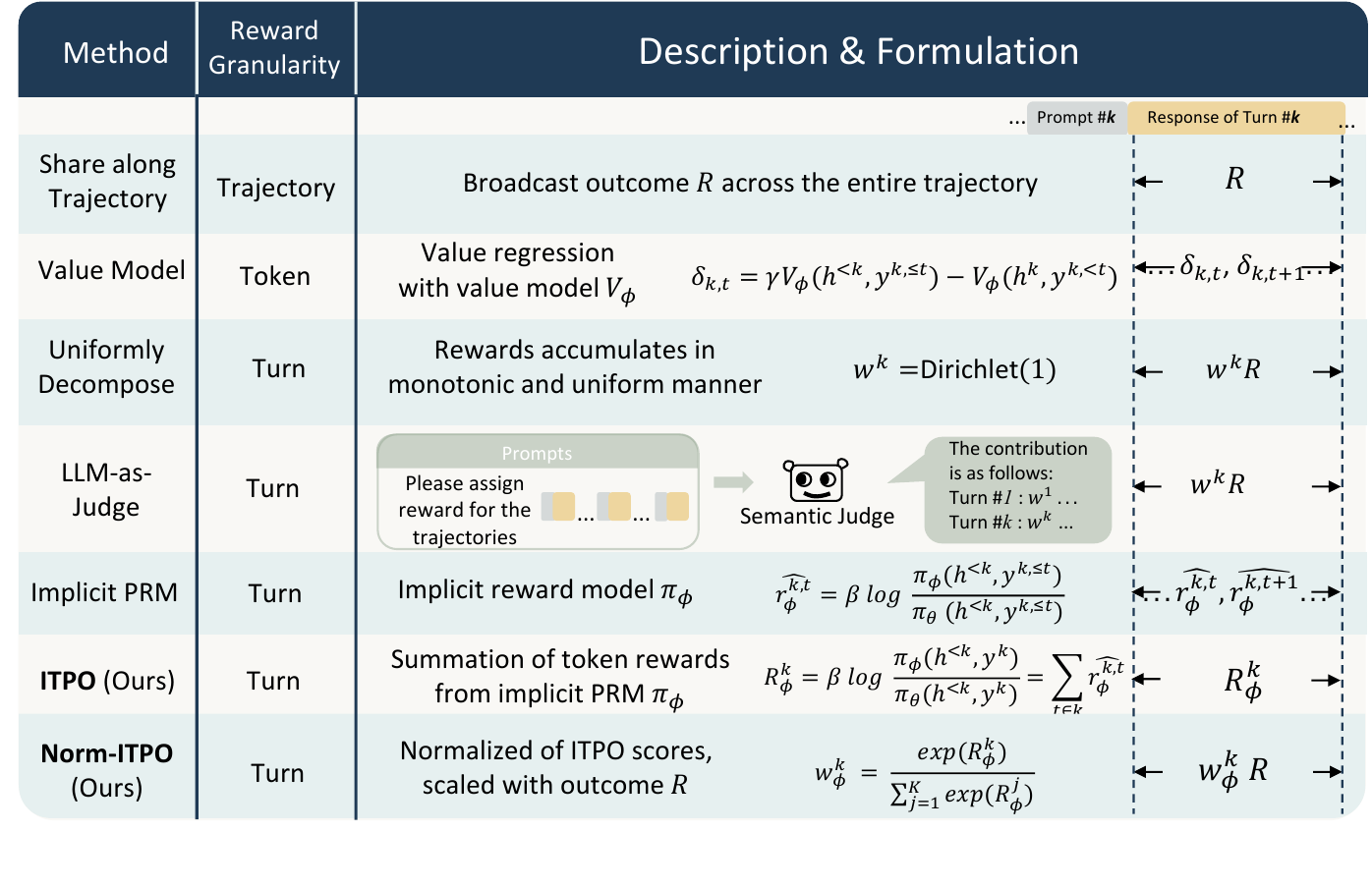}
    \vspace{-0.7cm}
    \caption{Comparison of the baselines adopted in the paper.}
    \label{fig:method_compare}
\end{figure}

The illustration of each baseline method is shown in Fig.~\ref{fig:method_compare}. The prompt of the LLM judge baseline in the three tasks are given below.

\begin{systembox}[title=System Prompt for LLM-as-a-Judge Baseline in Math Tutoring]
You are an expert Dialogue Analyst and Credit Assigner.
You will be reviewing a multi-turn collaboration between an **Assistant** and a **User**.

1. Input Context:
    - **Conversation History:** The step-by-step dialogue where the Assistant solves the problems or asks clarifying questions.
    - **User's Problem and the Solution:** The specific math problem and solution the user ultimately wants.

2. Objective:
    - Assign a **proportional reward value** (a decimal between 0.0 and 1.0) to each assistant's utterance (identified by utterance number) based on its contribution toward reaching the Solution of the Math Problem.
    - **The sum of all assigned reward values should equal 1.0**

---
Provided Information:

<|The Start of The Conversation|>
{conversation}
<|The End of The Conversation|>


<|The Start of The Target Problem|>
{target_problem}
<|The End of The Target Problem|>

<|The Start of The Ground Truth Solution|>
{target_solution}
<|The End of The Ground Truth Solution|>
---

## Please format your response as JSON with the following structure:

{{
    "thought": your brief thought,
    {Utterance_Demo},
}}

The utterance numbers should correspond to their order in the conversation.
Please annotate every utterance made by the assistant in the conversation, denoted "Utterance X by Assistant".
Please give a proportional reward value even if the utterance is the end of the conversation.

Double check if the JSON object is formatted correctly. Ensure that all fields are present and properly structured. \
Use " or """ to wrap up the thought and use single quotes inside the "thought" field to avoid JSON escape issues.
\end{systembox}

\begin{systembox}[title=Syetem Prompt for LLM-as-a-Judge Baseline in Doc Writing]

You are an expert Dialogue Analyst and Credit Assigner.
You will be reviewing a multi-turn collaboration between an **Assistant** and a **User**.
The Assistant's objective is to edit a document through conversation so that the final result matches the User's Target Document.

1. Input Context:
    - **Conversation History:** The step-by-step dialogue where the Assistant proposes edits or asks clarifying questions.
    - **User's Target Document:** The specific content and format the user ultimately wants.

2. Objective:
    - Assign a **proportional reward value** (a decimal between 0.0 and 1.0) to each assistant's utterance (identified by utterance number) based on its contribution toward reaching the User's Target Document.
    - **The sum of all assigned reward values should equal 1.0**

---
Provided Information:

<|The Start of The Conversation|>
{conversation}
<|The End of The Conversation|>

<|The Start of The Target Document|>
{target_document}
<|The End of The Target Document|>
---

## Please format your response as JSON with the following structure:

{{
    "thought": your brief thought,
    {Utterance_Demo},
}}

The utterance numbers should correspond to their order in the conversation.
Please annotate every utterance made by the assistant in the conversation, denoted "Utterance X by Assistant".
Please give a score even if the utterance is the end of the conversation.

Double check if the JSON object is formatted correctly. Ensure that all fields are present and properly structured. \
Use " or """ to wrap up the thought and use single quotes inside the "thought" field to avoid JSON escape issues.
\end{systembox}
\begin{systembox}[title=System Prompt for LLM-as-a-Judge Baseline in Medical Recommendation]
You are an expert Dialogue Analyst and Credit Assigner.
You will be reviewing a multi-turn collaboration between an **Assistant** and a **User**.

1. Input Context:
    - **Conversation History:** The step-by-step dialogue where the Assistant try to provide diagnosis and recommendation for patients.
    - **Patient's Symptoms and the Ground Truth:** The symptoms and corresponding diagnosis that the user needs.

2. Objective:
    - Assign a **proportional reward value** (a decimal between 0.0 and 1.0) to each assistant's utterance (identified by utterance number) based on its contribution towards the correct diagnosis and recommendation.
    - **The sum of all assigned reward values should equal 1.0**

---
Provided Information:
<|The Start of The Conversation|>
{conversation}
<|The End of The Conversation|>

<|The Start of The Symptoms|>
{symptom}
<|The End of The Symptoms|>

<|The Start of The Diagnosis|>
{diagnosis}
<|The End of The Diagnosis|>

<|The Start of The Recommendation|>
{recommendation}
<|The End of The Recommendation|>
---

## Please format your response as JSON with the following structure:

{{
    "thought": your brief thought,
    {Utterance_Demo},
}}

The utterance numbers should correspond to their order in the conversation.
Please annotate every utterance made by the assistant in the conversation, denoted "Utterance X by Assistant".
Please give a proportional reward value even if the utterance is the end of the conversation.

Double check if the JSON object is formatted correctly. Ensure that all fields are present and properly structured. \
Use " or """ to wrap up the thought and use single quotes inside the "thought" field to avoid JSON escape issues.
\end{systembox}

\subsubsection{\proj Implementation Details}
\label{app:itpo_implementation_details}
The hyper-parameters adopted by PRIME~\cite{cui2025process} baseline, \proj and Norm-\proj are listed in Table.~\ref{tab:hyper_parameter}, which is used across RLOO, GRPO and PPO advantage estimators. Both PRIME and \proj requires additional advantage information calculated from outcome reward $\mathbf{R}$ to converge, while Norm-ITPO does not require the additional information.

\begin{table}[h]
\centering
\resizebox{0.9\linewidth}{!}{\begin{tabular}{@{}c|ccc|ccc|ccc@{}}
\toprule
 & \multicolumn{3}{c|}{Math Tutoring} & \multicolumn{3}{c|}{Doc Writing} & \multicolumn{3}{c}{Medical Recommendation} \\ \midrule
 & $\beta$ & $\eta$ & $w(\mathcal{R}_{\phi})/w(\mathbf{R})$ & $\beta$ & $\eta$ & $w(\mathcal{R}_{\phi}/w(\mathbf{R})$ & $\beta$ & $\eta$ & $w(\mathcal{R}_{\phi})/w(\mathbf{R})$ \\
PRIME & 5e-2 & - & 5/5 & 5e-3 & - & 5/5 & 5e-3 & - & 5/5 \\
ITPO & 5e-2 & - & 5/5 & 5e-3 & - & 5/5 & 1e-2 & - & 5/5 \\
Norm-ITPO & 1e-2 & 0.4 & 5/0 & 5e-3 & 0.4 & 5/0 & 5e-3 & 0.4 & 5/0 \\ \bottomrule
\end{tabular}}
\caption{Hyperparameters of implicit process reward models. $w(\mathcal{R}_{\phi})/w(\mathbf{R})$ denotes the weight coefficient of advantage results from implicit reward $\mathcal{R}_{\phi}$ and outcome $\mathbf{R}$.}
\label{tab:hyper_parameter}
\end{table}

The algorithm of \proj is displayed in Algorithm.~\ref{alg:itpo}.
\begin{algorithm}[t]
    \caption{The \proj and Norm-\proj Framework}
    \label{alg:itpo}
    \begin{algorithmic}[1]
    \small
        \REQUIRE Simulator $U$, Policy $\pi_\theta$, Reference $\pi_{\text{ref}}$, Implicit PRM $\pi_\phi$
        \REQUIRE Hyperparameters: $\beta$, $\eta$, Learning rates $\eta_\theta, \eta_\phi$
        
        \FOR{iteration $iter = 1, \dots, N$}
            \STATE \textbf{Stage 1: Online Roll-out}
            \STATE Initialize buffer $\mathcal{D} \leftarrow \emptyset$
            \FOR{trajectory $i = 1, \dots, B$}
                \STATE Sample $\tau_i \sim \pi_\theta$ interactively with $U$
                \STATE Obtain outcome reward $\mathbf{R}^i \in [0,1]$
                \STATE Add $(\tau_i, \mathbf{R}^i)$ to $\mathcal{D}$
            \ENDFOR
            
            \STATE
            \STATE \textbf{Stage 2: Implicit PRM Update}
            \STATE Compute implicit loss $\mathcal{L}_{\text{PRM}}(\phi)$ on $\mathcal{D}$ \COMMENT{Eq.~\ref{eq:prm_loss}}
            \STATE Update $\phi \leftarrow \phi - \eta_\phi \nabla_\phi \mathcal{L}_{\text{PRM}}(\phi)$
            \STATE Compute token-level $r_\phi(y^{i,k,t})$ with updated $\pi_\phi$ \COMMENT{Eq.~\ref{eq:reward_def}}
            
            \STATE
            \STATE \textbf{Stage 3: Reward Assignment}
            \FOR{each trajectory $(\tau_i, \mathbf{R}^i)$ in $\mathcal{D}$}
                \STATE Turn-level evidence $\mathcal{R}_\phi^{i,k} \leftarrow \sum_{t} r_\phi^{i,k,t})$ \COMMENT{Eq.~\ref{eq:itpo_score}}
                \IF{\texttt{UseNorm} is \textbf{True}}
                    \STATE \textbf{(A) Norm-ITPO}: 
                    \STATE $w_{\phi}^{i,k} \leftarrow \text{Normalization} (\mathcal{R}_\phi)$
                    \COMMENT{Eq.~\ref{eq:norm_weights}}
                    \STATE Assign $\tilde{\mathcal{R}}_{\phi}^{i,k} \leftarrow w^{i,k}_{\phi} \cdot \mathbf{R}^i$ \COMMENT{Eq.~\ref{eq:norm_reward}}
                \ELSE
                    \STATE \textbf{(B) ITPO}: 
                    \STATE Assign $\mathcal{R}_{\phi}^{i,k}$ \COMMENT{\scriptsize Raw turn-wise process reward}
                \ENDIF
            \ENDFOR
            
            \STATE
            \STATE \textbf{Stage 4: Policy Optimization}
            \STATE Estimate $\hat{A}^{i,k}$ with turn-wise rewards \COMMENT{\scriptsize e.g., GAE, GRPO, RLOO}
            \STATE Update $\pi_\theta \leftarrow \theta - \eta_\theta \nabla_\theta \mathcal{L}_{\text{PPO}}(\theta)$
        \ENDFOR
    \end{algorithmic}
\end{algorithm}

\subsection{More Experiment Results}

\subsubsection{Training Curves}
\begin{figure*}[h] 
    \centering
    \begin{subfigure}[b]{0.33\textwidth} 
        \includegraphics[width=\linewidth]{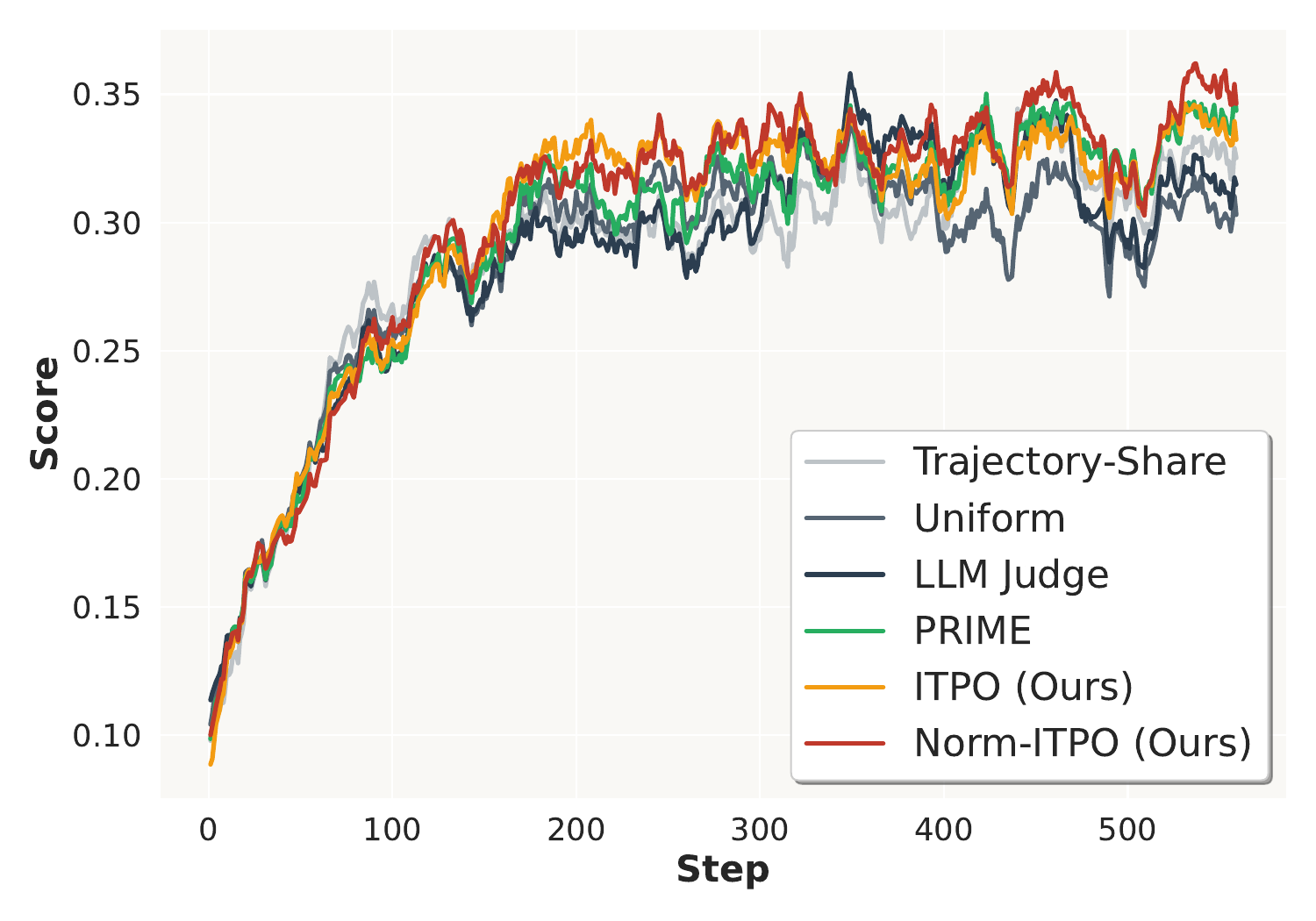} 
        \vspace{-0.7cm}
        \caption{Math tutoring.}
        \label{fig:math_grpo}
    \end{subfigure}
    \hfill 
    \begin{subfigure}[b]{0.33\textwidth}
        \includegraphics[width=\linewidth]{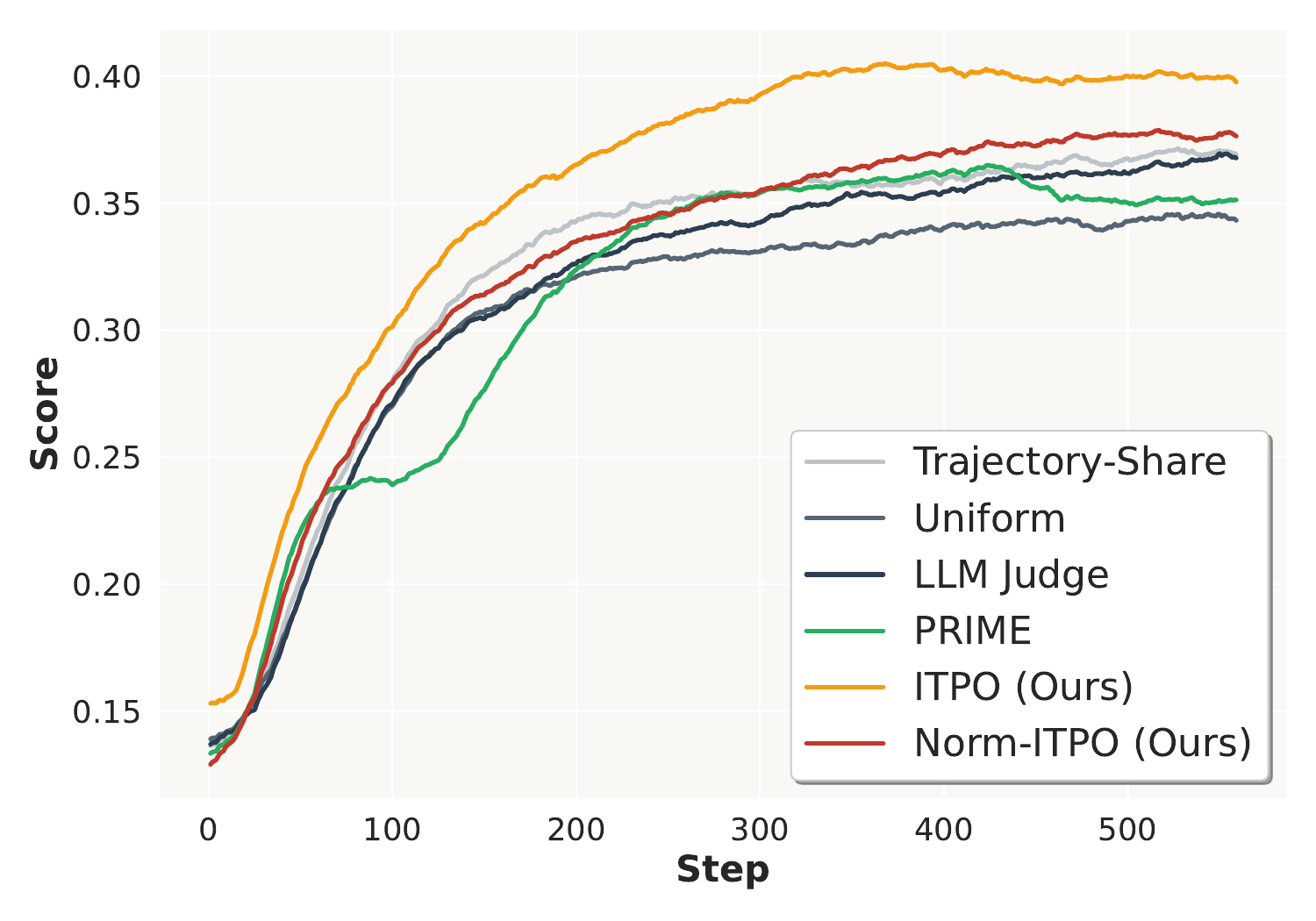}
         \vspace{-0.7cm}
        \caption{Document writing.}
        \label{fig:document_grpo}
    \end{subfigure}
    \hfill 
    \begin{subfigure}[b]{0.33\textwidth}
        \includegraphics[width=\linewidth]{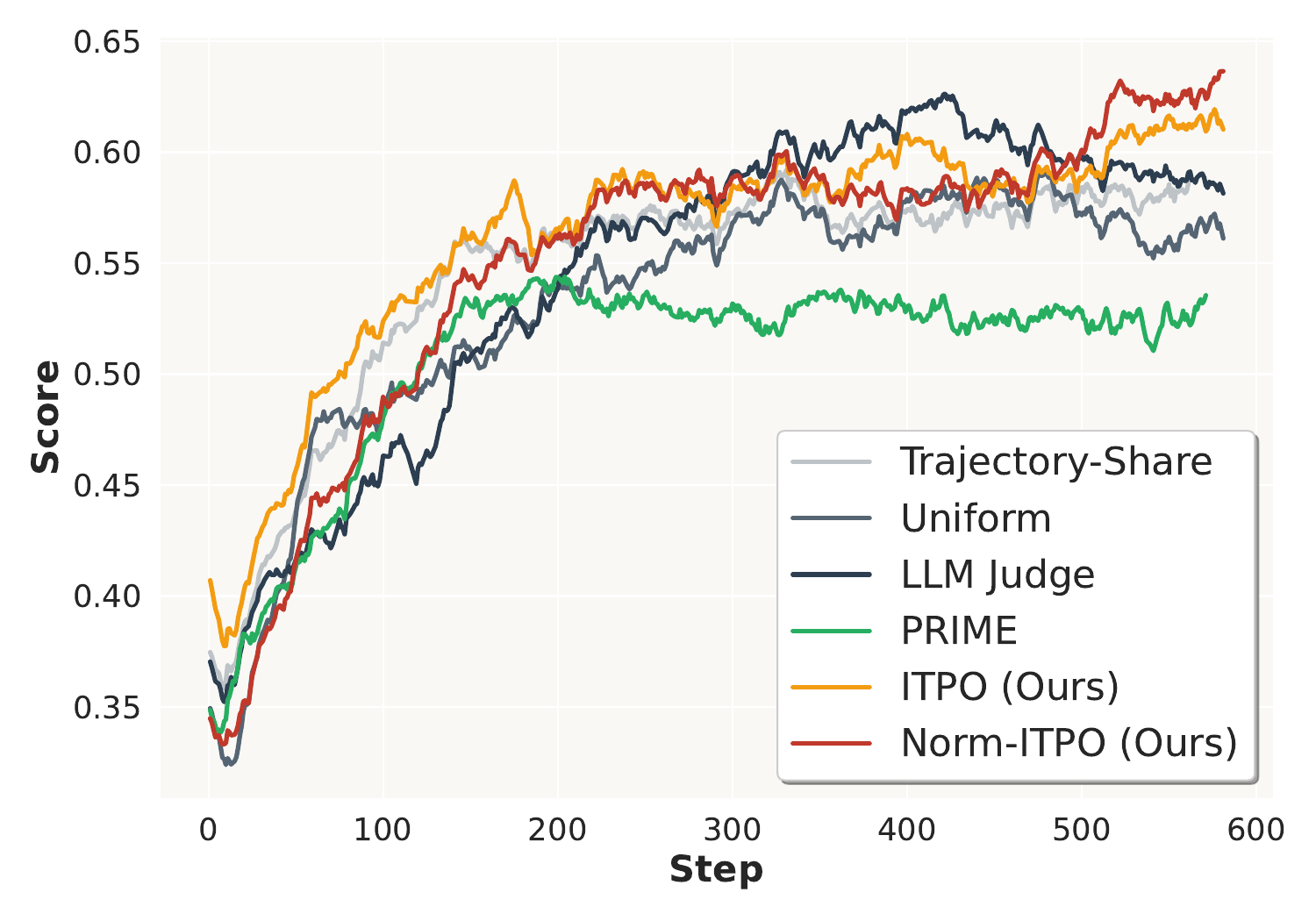}
         \vspace{-0.7cm}
        \caption{Medical recommendation.}
        \label{fig:medical_grpo}
    \end{subfigure}
    \caption{The training curve of reward attribution methods with the GRPO advantage estimator.}
    \label{fig:training_curve_grpo}
\end{figure*}

The training dynamics of Qwen2.5-3B-Instruct~\cite{yang2025qwen2} with GRPO~\cite{shao2024deepseekmath} is shown in Fig.~\ref{fig:training_curve_grpo}.

\subsubsection{Stability and Consistency Analysis}
\begin{figure}[h]
     \centering
     \begin{subfigure}[b]{0.24\textwidth}
         \centering
         \includegraphics[width=\textwidth]{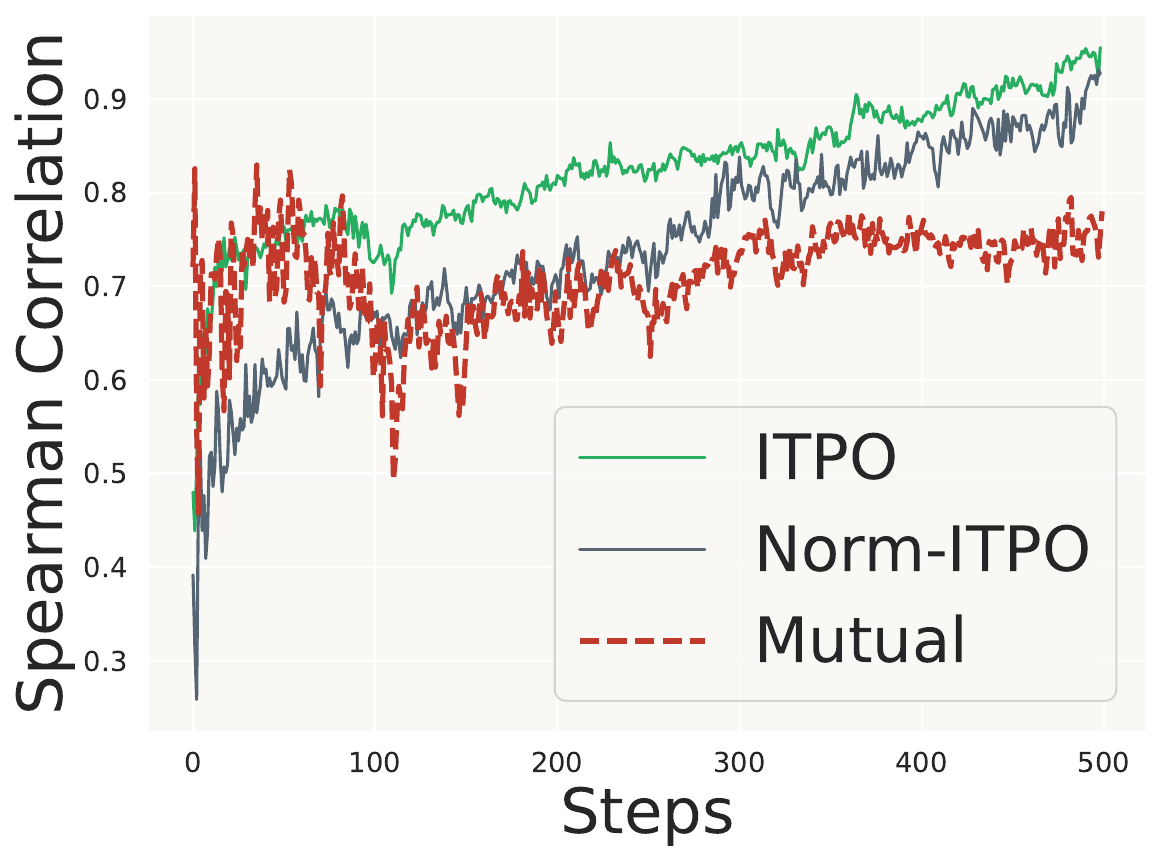}
         \caption{Convergence \& Stability (Math Tutoring).}
     \end{subfigure}
     \hfill
     \begin{subfigure}[b]{0.24\textwidth}
         \centering
         \includegraphics[width=\textwidth]{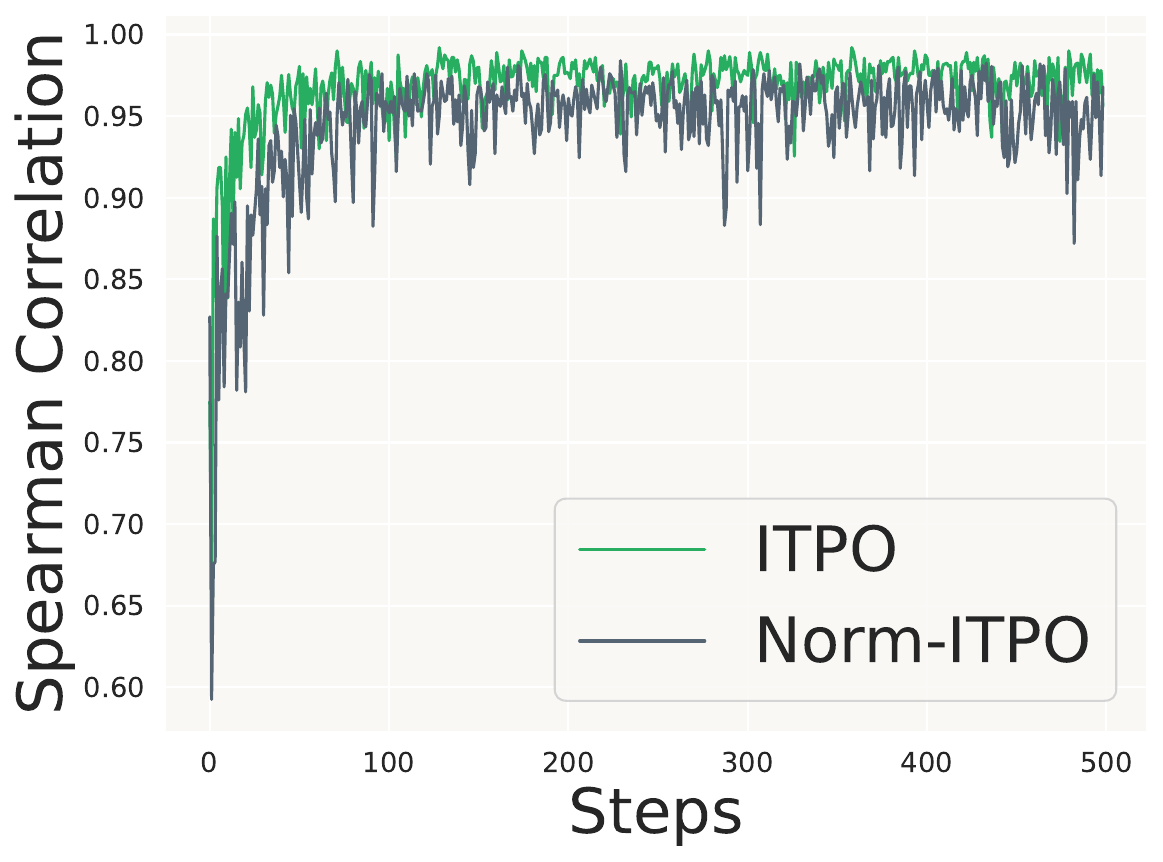}
         \caption{Self-Consistency (Math Tutoring).}
     \end{subfigure}
     \hfill
     \begin{subfigure}[b]{0.24\textwidth}
         \centering
         \includegraphics[width=\textwidth]{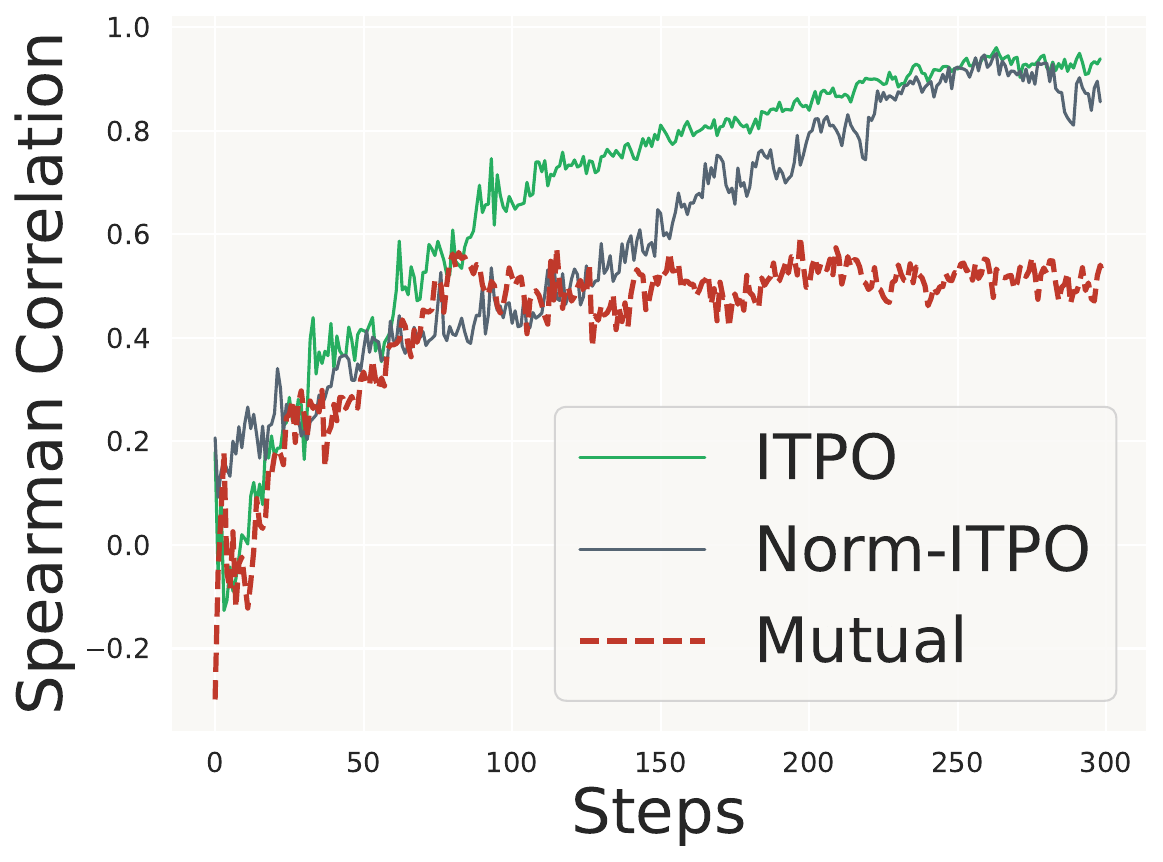}
         \caption{Convergence \& Stability (Medical Rec)}
     \end{subfigure}
     \hfill
     \begin{subfigure}[b]{0.24\textwidth}
         \centering
         \includegraphics[width=\textwidth]{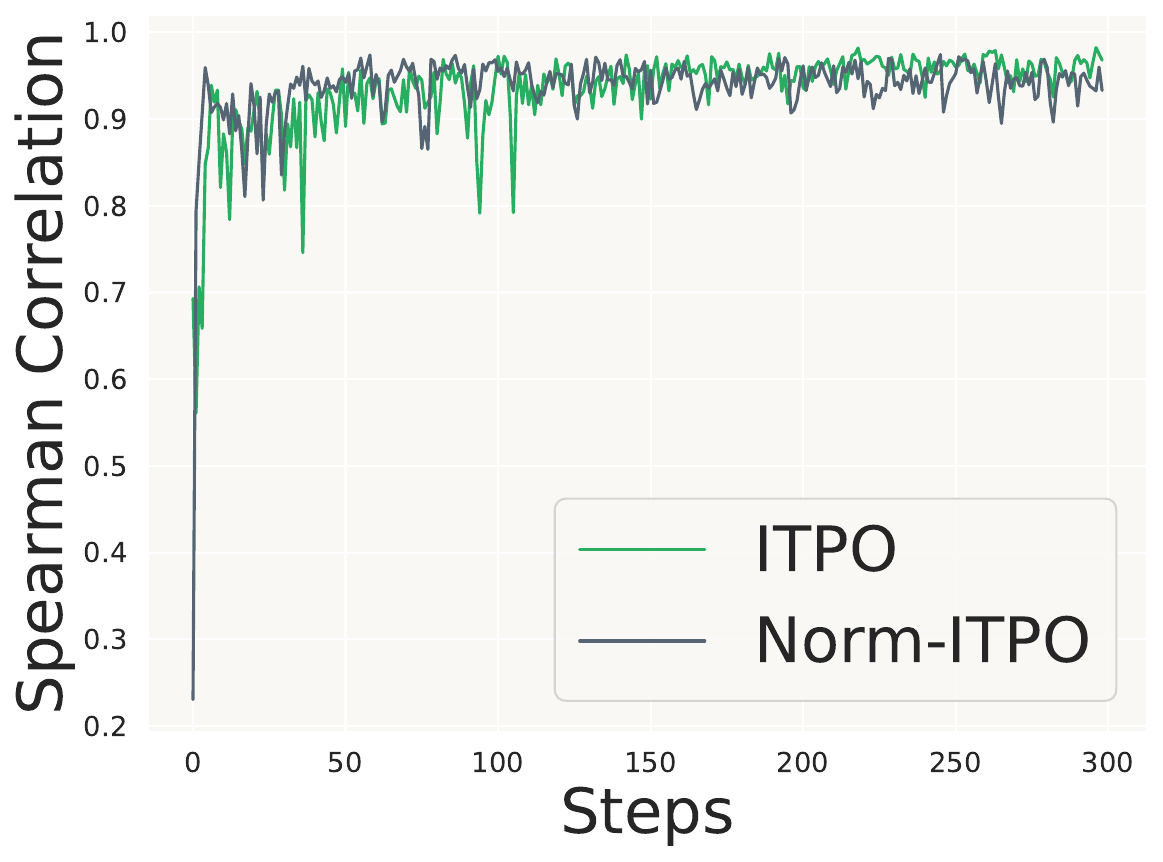}
         \caption{Self-Consistency (Medical Rec)}
     \end{subfigure}
     
     \caption{Stability and Consistency on the Math Tutoring and Medical Recommendation tasks.}
     \label{fig:stable_consistency_doc_medical}
\end{figure}
For the stability and consistency analysis on Doc Writing and Medical Recommendation, refer to Fig.~\ref{fig:stable_consistency_doc_medical}.

\subsubsection{Trajectory-Level Implicit Reward Analysis}
\begin{figure}[htbp]
    \centering
    \begin{subfigure}[b]{0.3\textwidth}
        \centering
        \includegraphics[width=\textwidth]{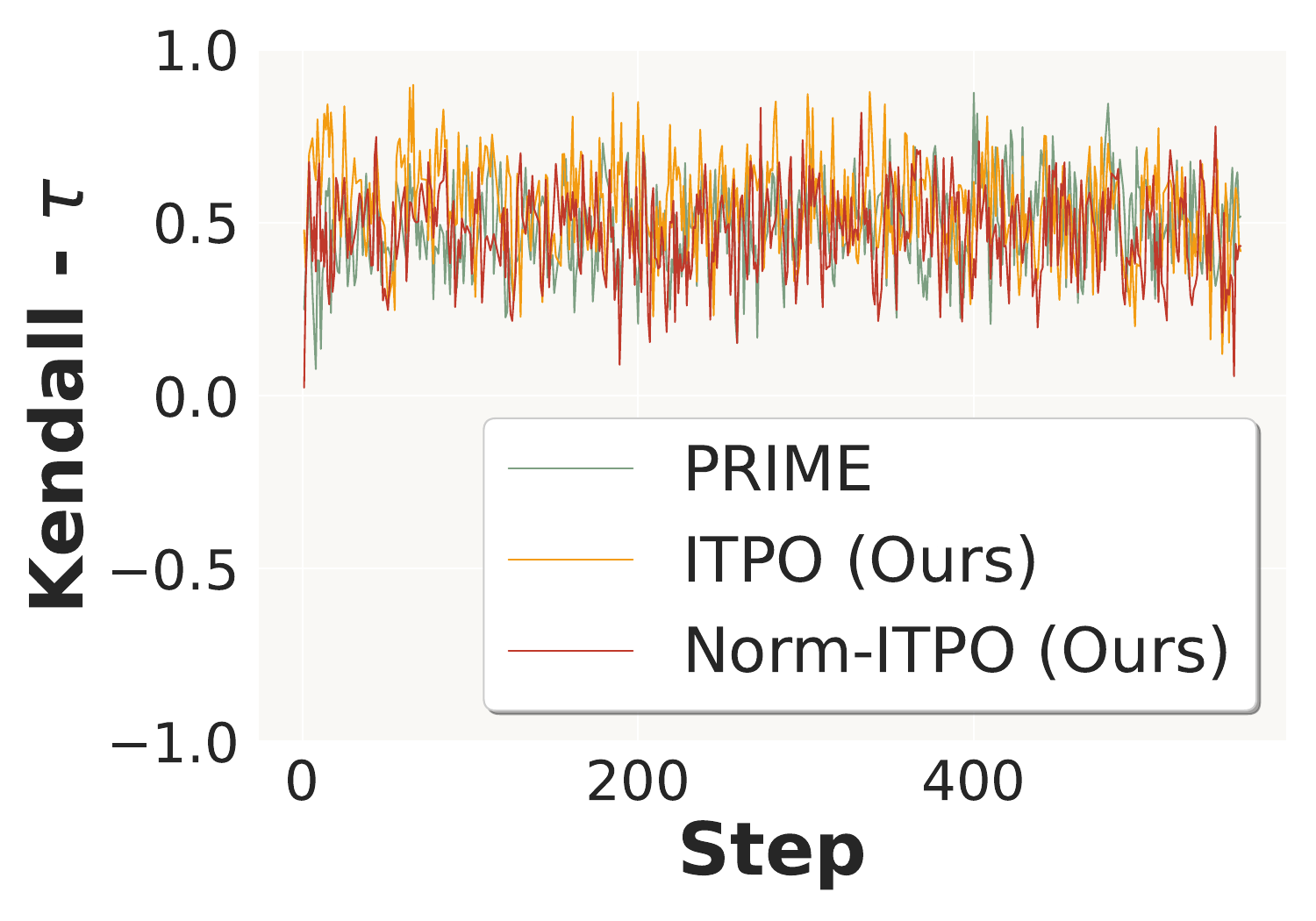} 
        \caption{Math Tutoring.}
    \end{subfigure}
    \hfill 
    \begin{subfigure}[b]{0.3\textwidth}
        \centering
        \includegraphics[width=\textwidth]{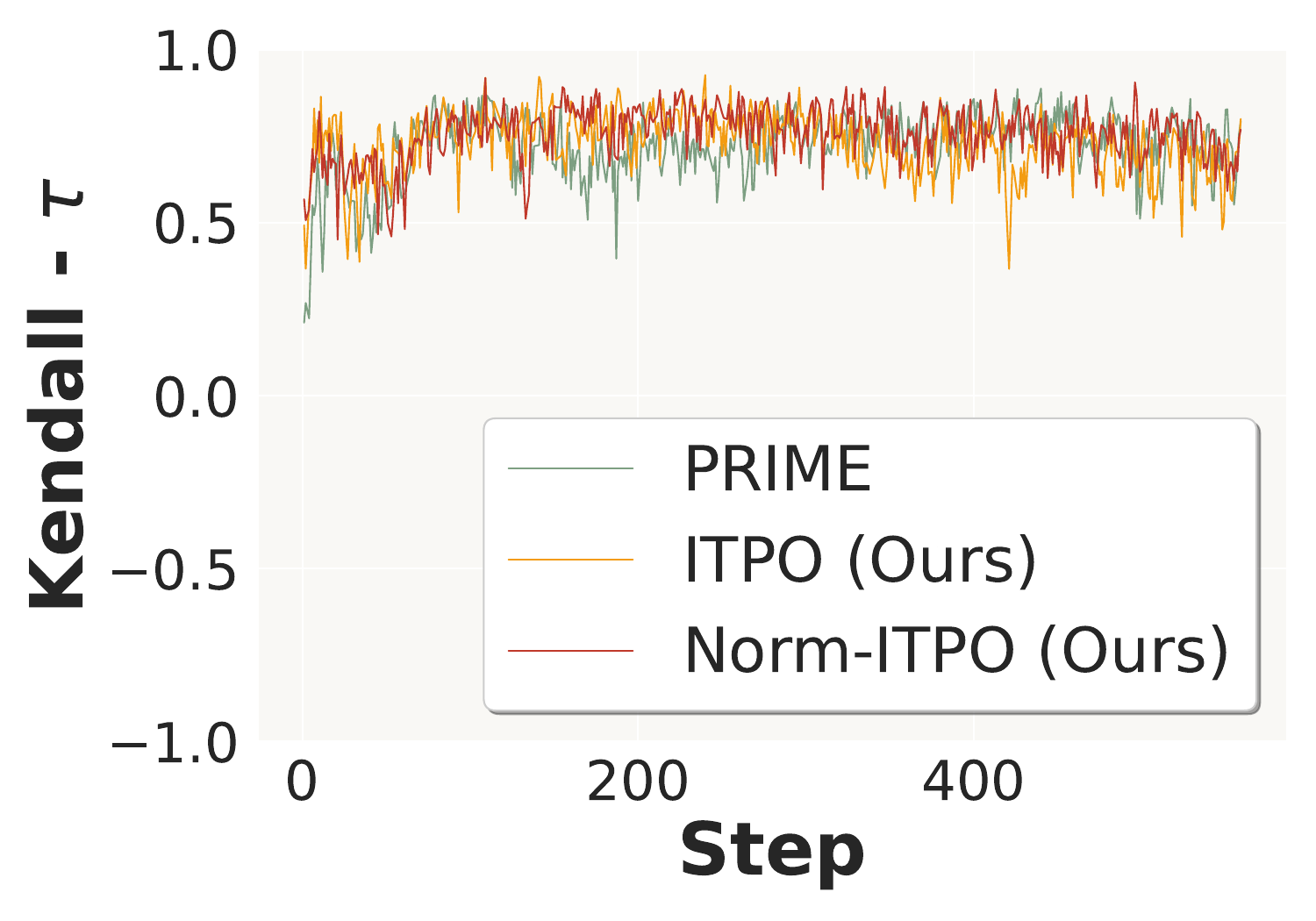}
        \caption{Doc Writing.}
    \end{subfigure}
    \hfill
    \begin{subfigure}[b]{0.3\textwidth}
        \centering
        \includegraphics[width=\textwidth]{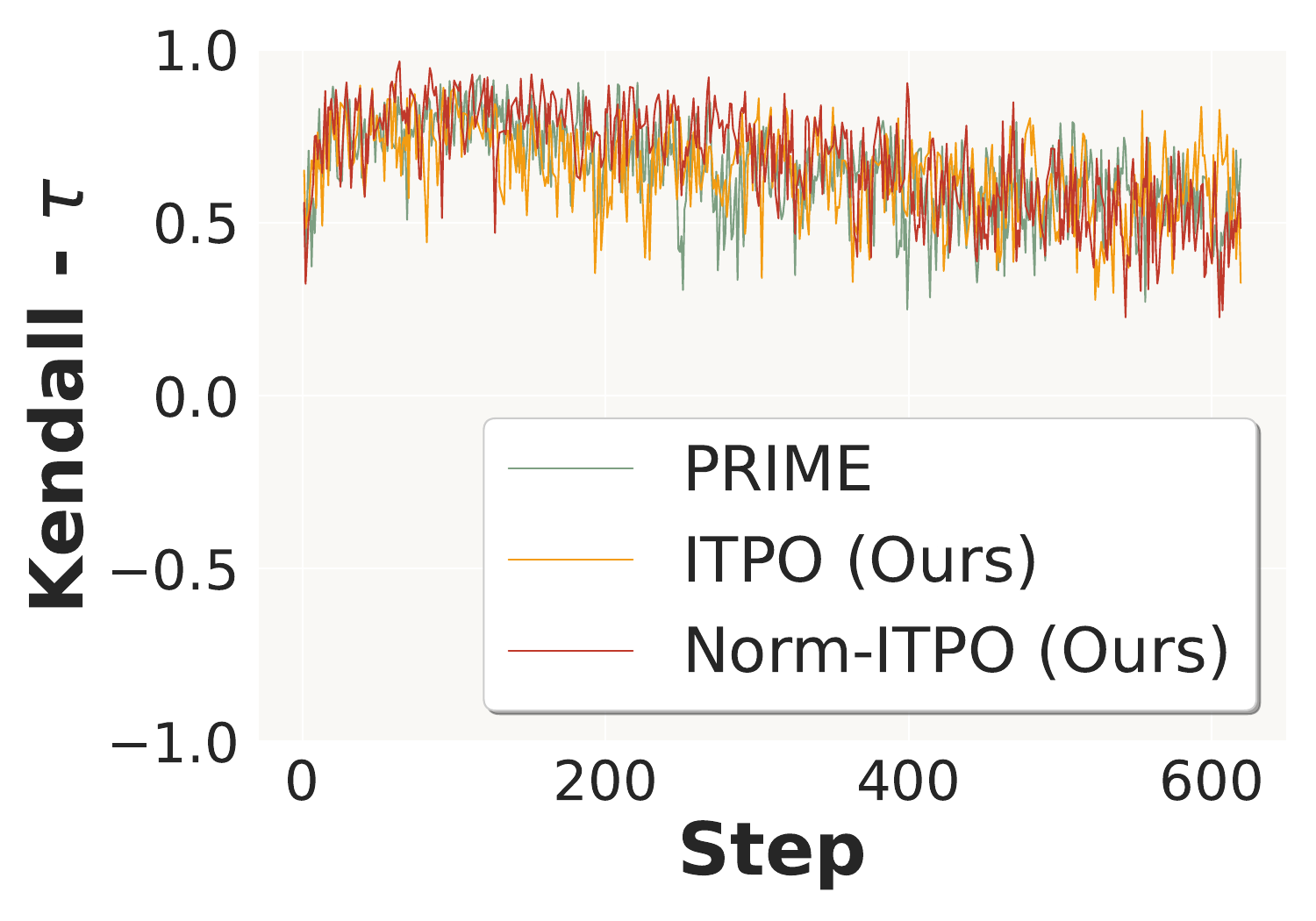}
        \caption{Medical Recommendation.}
    \end{subfigure}
    \caption{Kendall-$\tau$ between PRM and outcome reward w/ GRPO.}
    \label{fig:effectiveness_rm_grpo}
\end{figure}
For the trajectory-level implicit reward prediction evaluation with the GRPO advantage function, see Fig.~\ref{fig:effectiveness_rm_grpo}.

\subsection{More Related Works}
\label{app:more_related_works}
\subsection{Implicit Process Reward Models}
Implicit PRMs~\cite{yuan2024free, cui2025process, rafailov2024r} derive fine-grained token-level process rewards solely from verifiable outcome supervision. Specifically, these methods formulate the token-level process reward as the log-likelihood ratio between the reward model and the reference model, allowing the gradient from sparse outcome supervision to implicitly propagate credit to individual tokens. By eliminating the need for human process annotations, implicit PRMs offer a scalable alternative to explicit PRMs. 
Recent theoretical analysis~\cite{razin2025your} suggests that while the token-level implicit process reward can effectively fit training distributions as explicit PRMs, it often suffers from poor generalization. The high variance inherent in assigning dense token-level credit from sparse outcomes creates a noisy supervision signal, highlighting the need for stable improvements. Under the multi-turn settings, such variance is further exacerbated by the stochastic user dynamics; meanwhile, the fine-grained supervision on isolated tokens lacks semantic interpretability.

\textbf{User Simulation for Evaluation and Training}
User simulators have become pivotal for studying systems involving human interaction~\cite{guo2024embodied, sun2024parrot}, widely applied in social science~\cite{tang2025gensim}, economic modeling~\cite{horton2023large}, and debate~\cite{du2023improving}. 
Recent evaluations~\cite{dou2025simulatorarena, naous2025flipping} demonstrate that LLMs conditioned on detailed user profiles can closely align with human judgments and serve as reliable stand-ins for real users, enabling the training and evaluation of policies in a controlled environment.

\clearpage
\subsection{Semantic Interpretability}
\label{app:sematic_interpretability}
See the examples in the next page.

\newpage
\begin{figure}[h]
    \centering
    \includegraphics[width=0.99\linewidth]{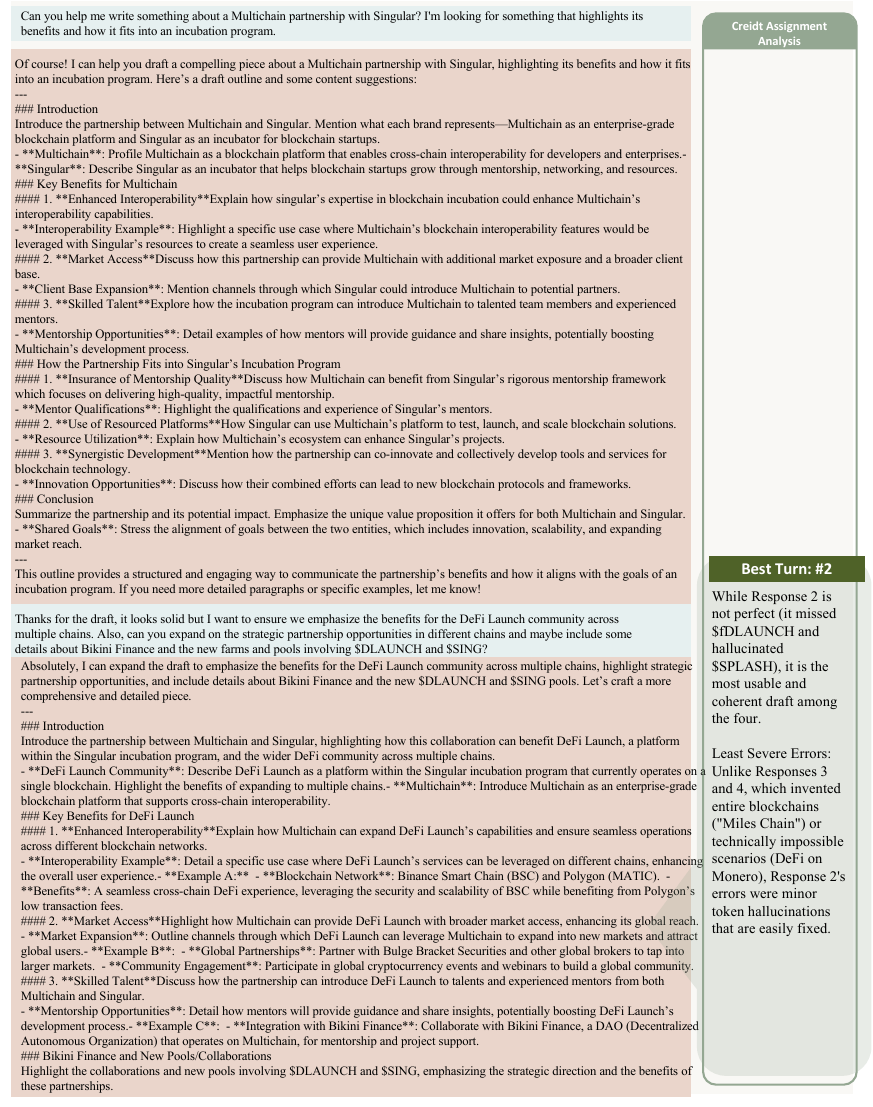}
    \caption{Part \#1 of low-performance trajectory \#1 in Example \#1}
    \label{fig:e1_low1_1}
\end{figure}

\begin{figure}
    \centering
    \includegraphics[width=0.99\linewidth]{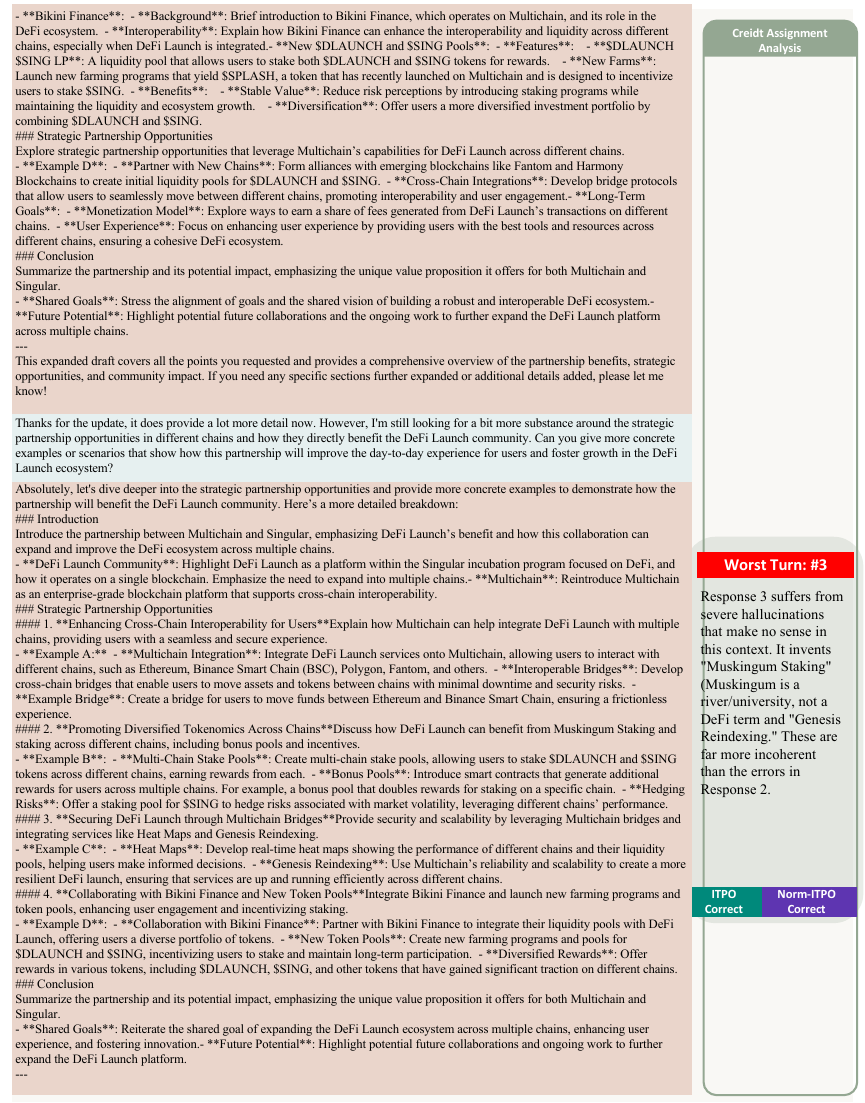}
    \caption{Part \#2 of low-performance trajectory \#1 in Example \#1}
    \label{fig:e1_low1_2}
\end{figure}

\begin{figure}
    \centering
    \includegraphics[width=0.99\linewidth]{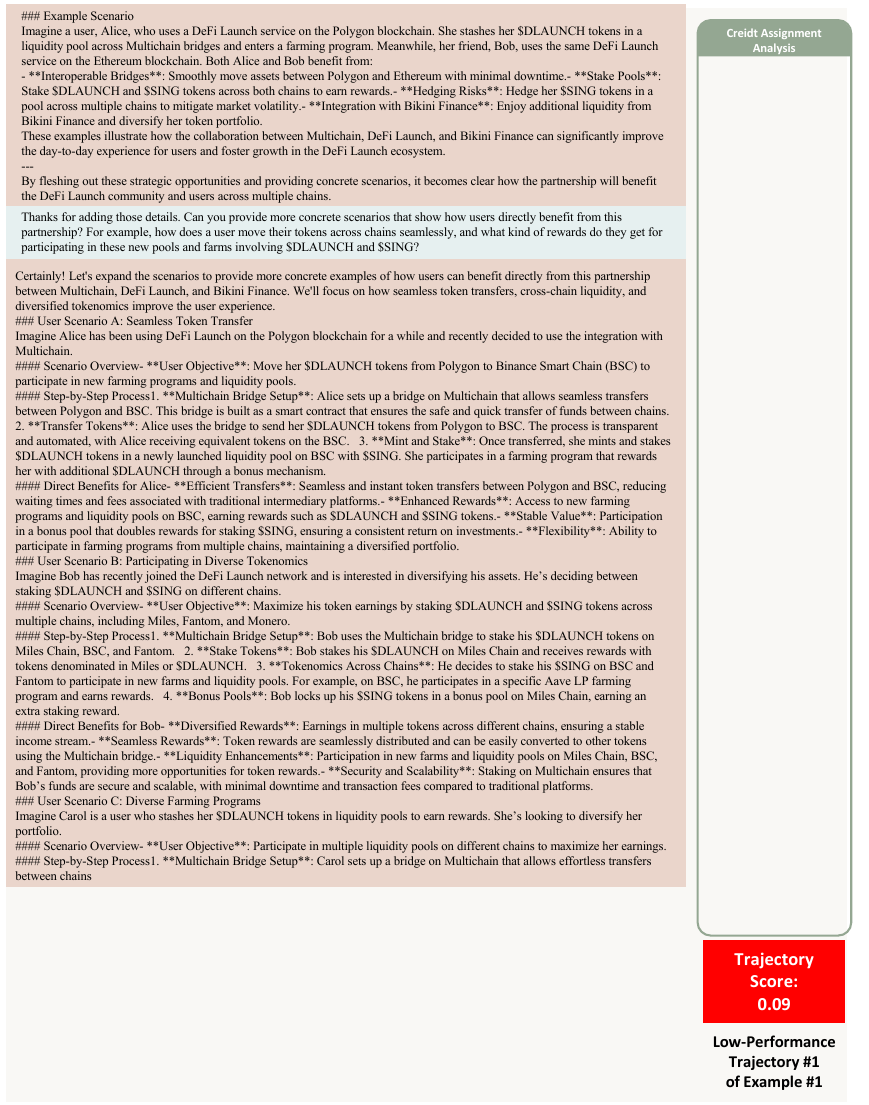}
    \caption{Part \#3 of low-performance trajectory in Example \#1}
    \label{fig:e1_low1_3}
\end{figure}

\begin{figure}
    \centering
    \includegraphics[width=0.99\linewidth]{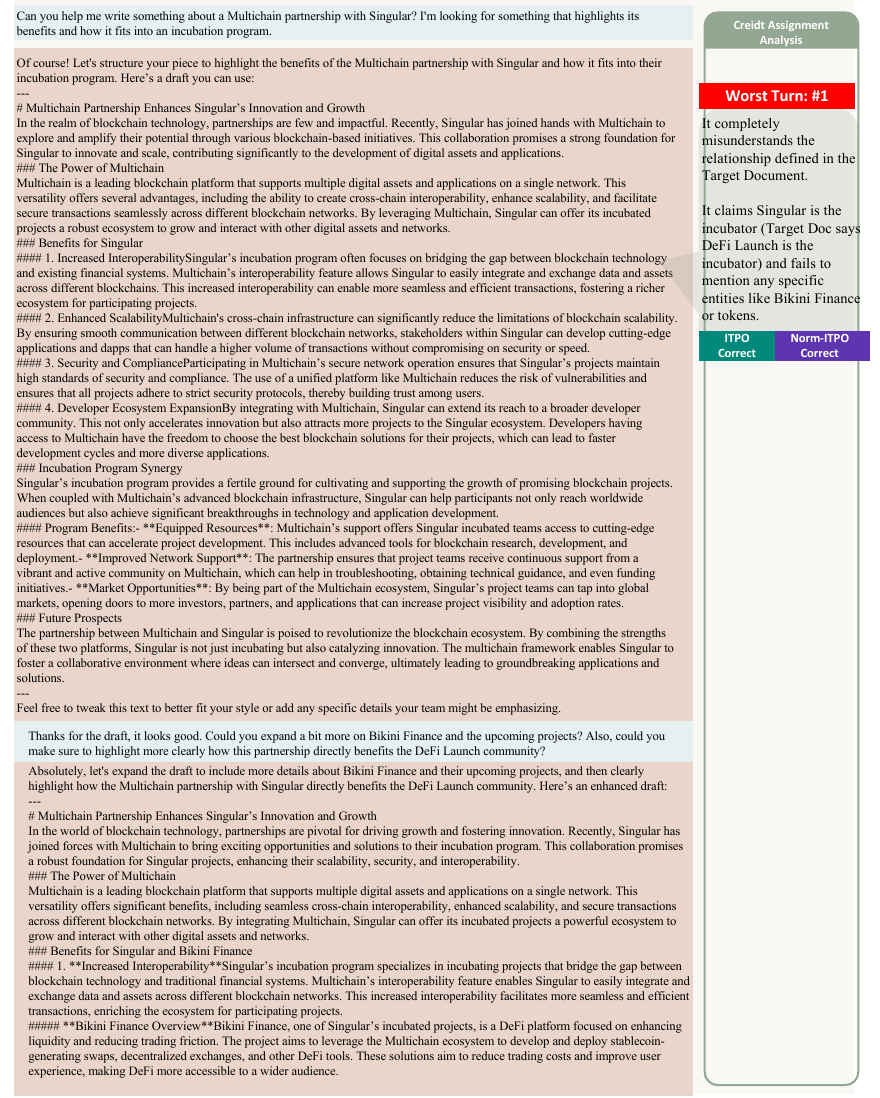}
    \caption{Part \#1 of low-performance trajectory \#2 in Example \#1}
    \label{fig:e1_low2_1}
\end{figure}

\begin{figure}
    \centering
    \includegraphics[width=0.99\linewidth]{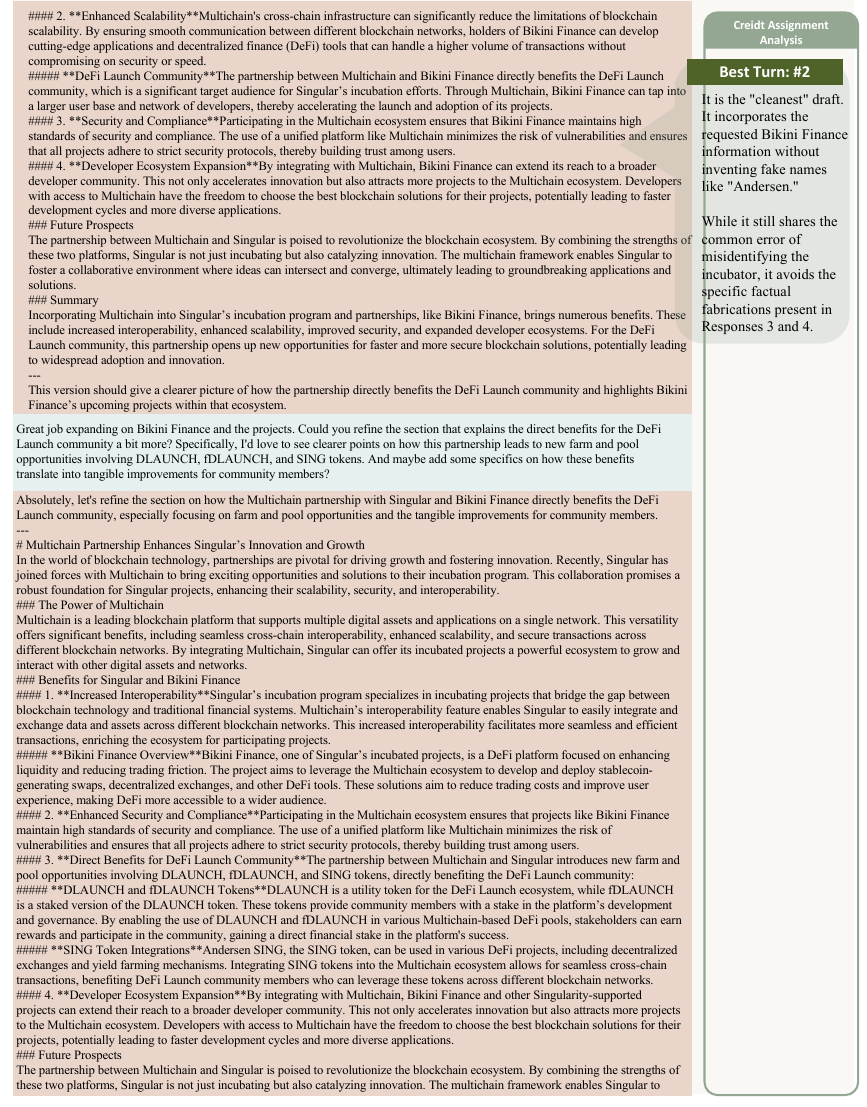}
    \caption{Part \#2 of low-performance trajectory \#2 in Example \#1}
    \label{fig:e1_low2_2}
\end{figure}

\begin{figure}
    \centering
    \includegraphics[width=0.99\linewidth]{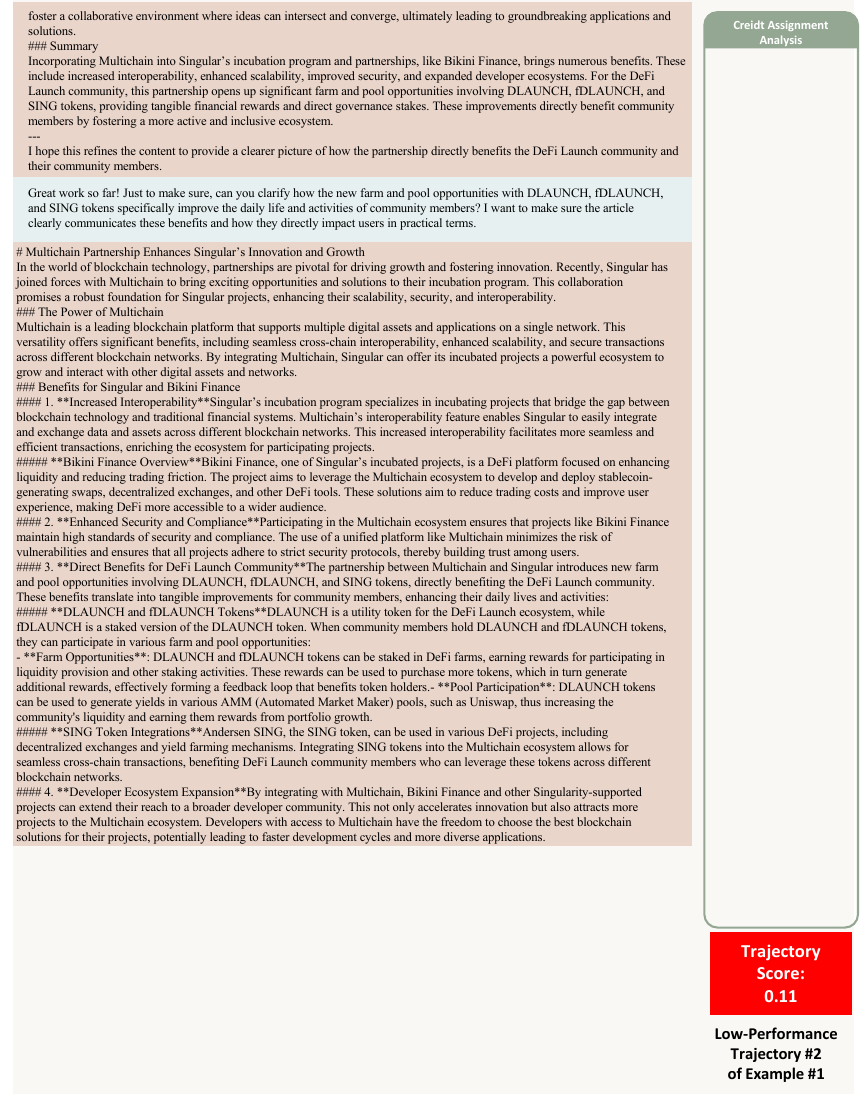}
    \caption{Part \#3 of low-performance trajectory \#2 in Example \#1}
    \label{fig:e1_low2_3}
\end{figure}

\begin{figure}
    \centering
    \includegraphics[width=0.99\linewidth]{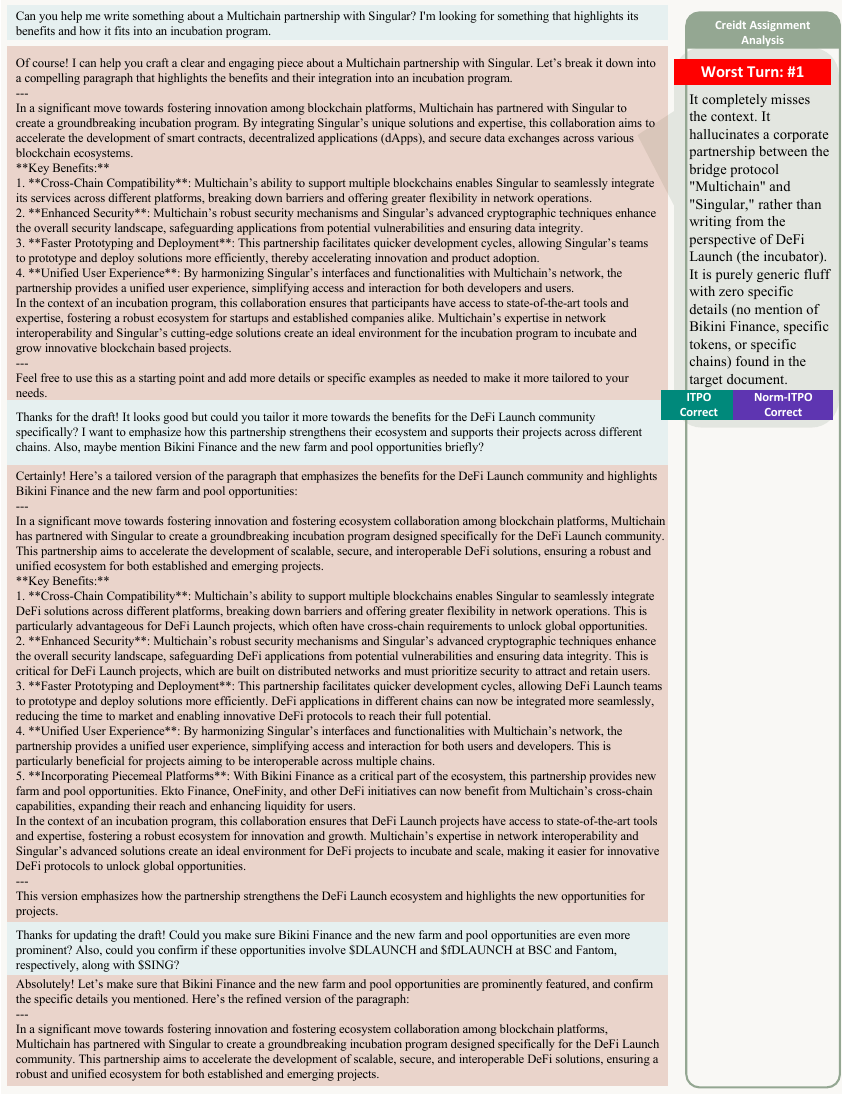}
    \caption{Part \#1 of high-performance trajectory \#1 in Example \#1}
    \label{fig:e1_high1_1}
\end{figure}

\begin{figure}
    \centering
    \includegraphics[width=0.99\linewidth]{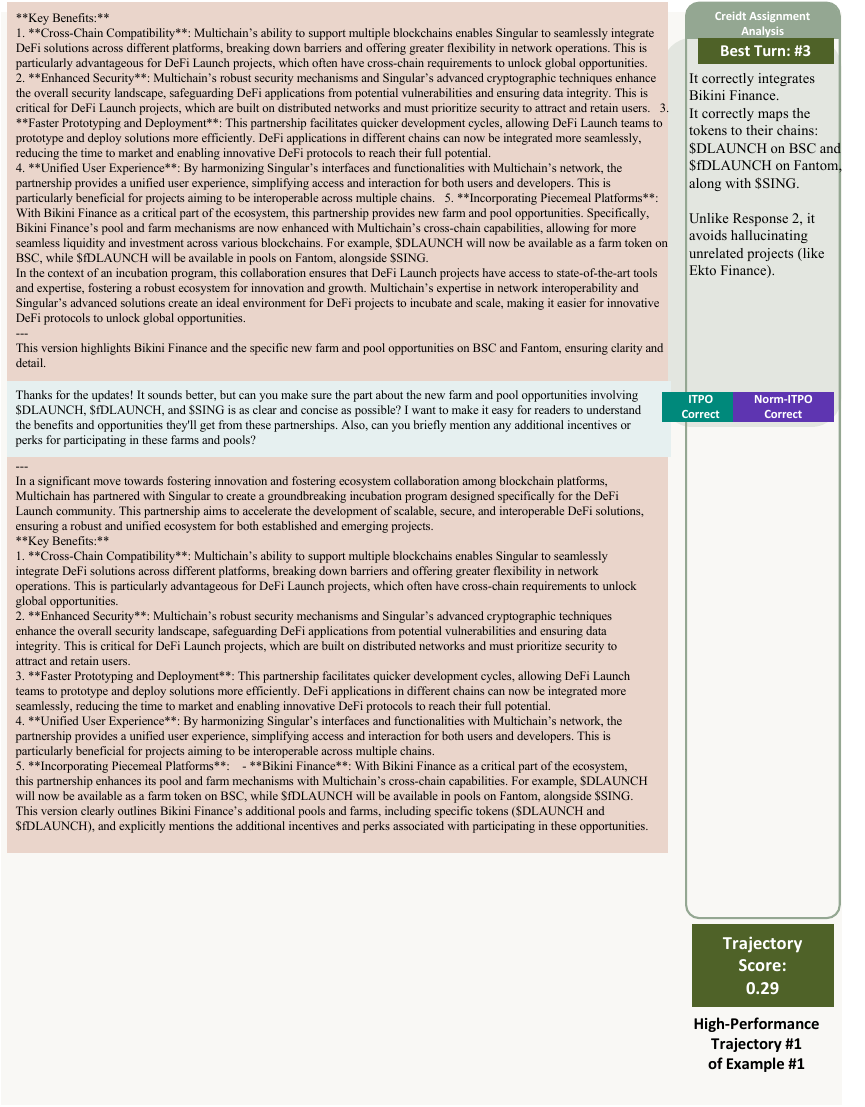}
    \caption{Part \#2 of high-performance trajectory \#1 in Example \#1}
    \label{fig:e1_high1_2}
\end{figure}

\begin{figure}
    \centering
    \includegraphics[width=0.99\linewidth]{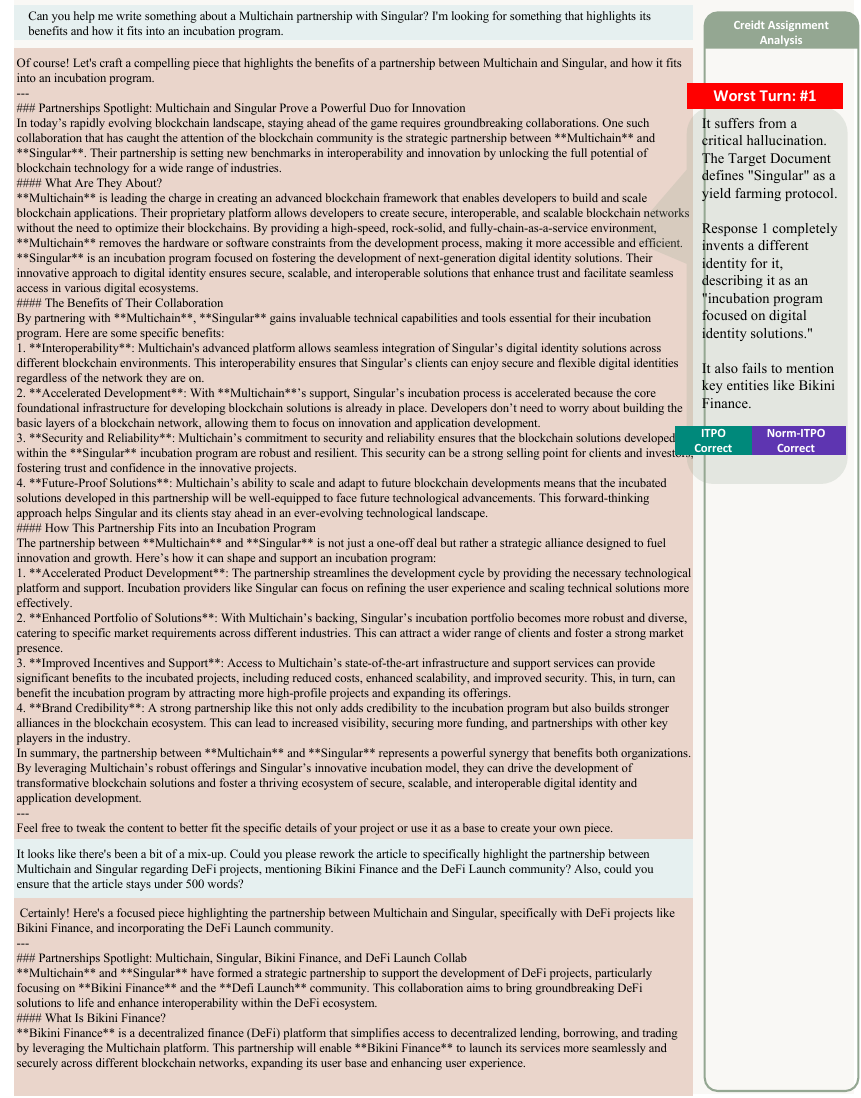}
    \caption{Part \#1 of high-performance trajectory \#2 in Example \#1}
    \label{fig:e1_high2_1}
\end{figure}

\begin{figure}
    \centering
    \includegraphics[width=0.99\linewidth]{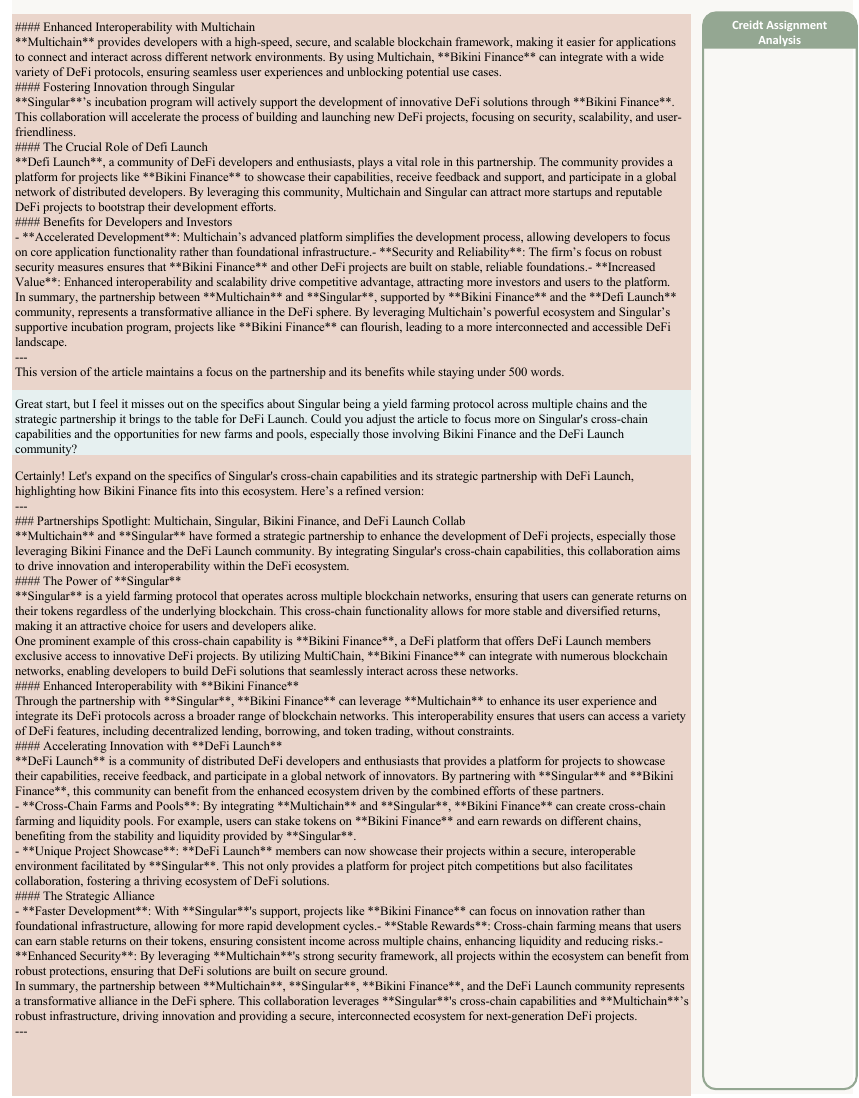}
    \caption{Part \#2 of high-performance trajectory \#2 in Example \#1}
    \label{fig:e1_high2_2}
\end{figure}

\begin{figure}
    \centering
    \includegraphics[width=0.99\linewidth]{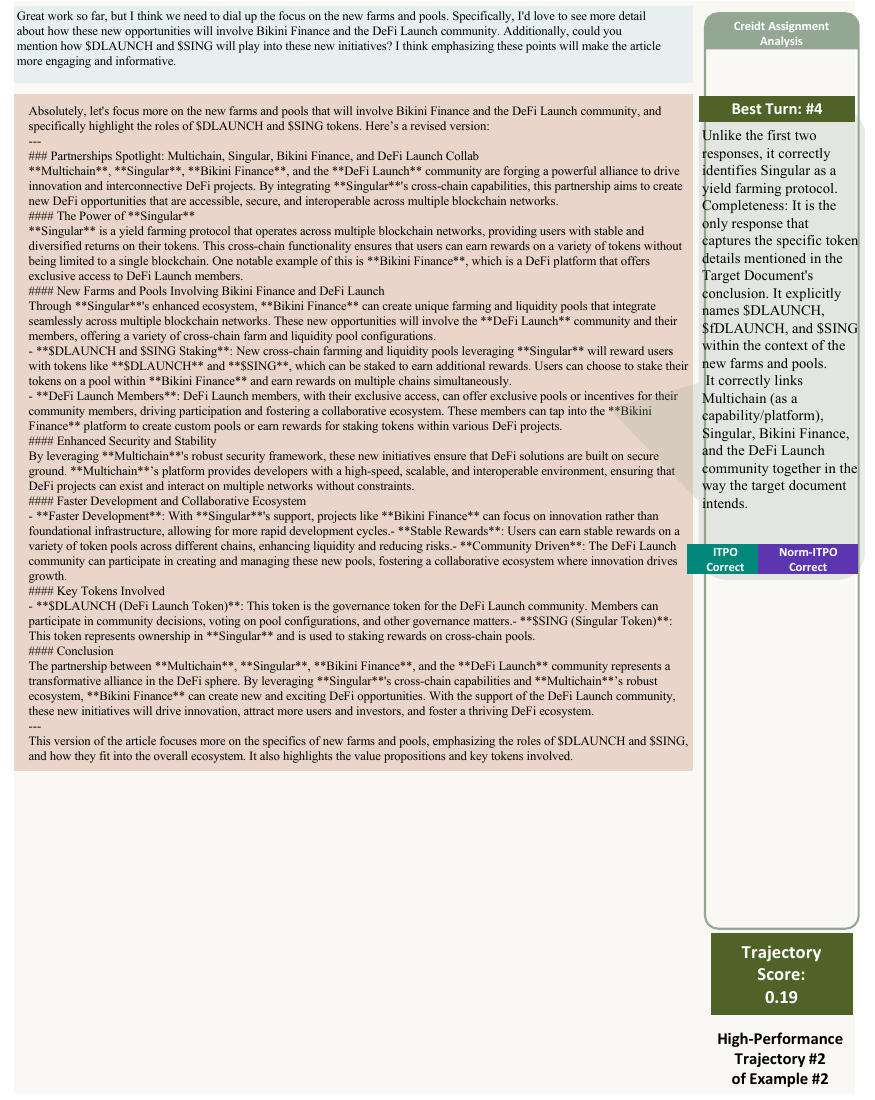}
    \caption{Part \#3 of high-performance trajectory \#2 in Example \#1}
    \label{fig:e1_high2_3}
\end{figure}

\clearpage

Above are the first $4$ examples with same initial prompts for turn-wise reward allocation interpretability analysis. For the rest $28$ examples, please refer to supplementary materials.

\end{document}